\documentclass[runningheads]{llncs}

\usepackage{eccv}

\usepackage{eccvabbrv}

\usepackage{graphicx}
\usepackage{booktabs}

\usepackage[accsupp]{axessibility}  

\usepackage{hyperref}

\usepackage{orcidlink}

\usepackage{tabularx}
\usepackage[table]{xcolor}
\usepackage{float}
\usepackage{color}

\begin{document}

\title{HOIMask: Towards Generative Masked Modeling for Human Object Interaction Generation} 

\titlerunning{HOIMask: Generative Masked Modeling for Human Object Interaction}

\author{Yihong Ji\inst{1, 2}\thanks{Corresponding author.} \and  
Jinsong Zhang\inst{3} \and     
He Hu\inst{1, 2} \and
Hongbo Xu\inst{2}}

\authorrunning{Y. Ji et al.}

\institute{College of Computer Science and Software Engineering, Shenzhen University \and 
Guangdong Laboratory of Artificial Intelligence and Digital Economy (SZ) \and
Fuzhou University}

\maketitle

\begin{center}
     \centering
     \includegraphics[width=\linewidth]{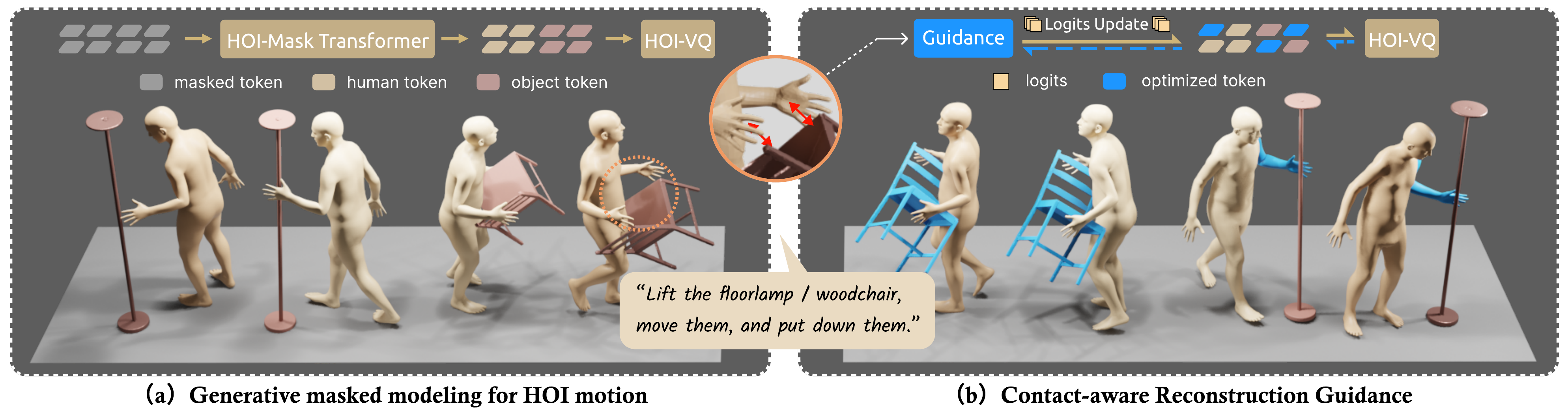}
     \captionof{figure}{HOIMask is a generative masked model to synthesis high fidelity Human-Object Interaction (HOI) motion from text description and object geometry. (a) Achieving accurate spatial and temporal coordination  with our proposed HOI VQ-VAE and HOI-Mask Transformer, and (b) guiding the model to generate more rational interaction through Contact-aware Reconstruction Guidance.}
 \end{center}%

\begin{abstract}
  Diffusion-based methods have dominated the HOI generation, as they enable critical contact fusions or signals to guide the diffusion process. However, they often result in high artifacts and unstable interaction quality due to error accumulation during iterative denoising. In this work, we propose HOIMask, the first generative masked framework for modeling HOI motion in discrete space. HOIMask first encodes both motion sequences and contact-aware signals into discrete 2D human and object token maps via HOI Vector Quantization (VQ), preserving fine-grained spatial-temporal structure beyond conventional 1D representations. On this basis, a generative masked modeling framework is employed to jointly capture human-object interaction dynamics, leveraging a transformer architecture designed to model complex spatial-temporal and interaction dependencies. To generate more coherent and physically plausible motions, we further introduce a novel contact-aware reconstruction guidance in discrete space during inference, which fuses contact signals to optimize HOI tokens that forces the generated motion with higher spatio-temporal consistency. With craftily designed motion interaction tokens, dedicated architecture and guidance strategy, HOIMask outperforms state-of-the-art diffusion-based methods, generating more realistic and semantically aligned HOI motions. Please refer to our \href{https://jyhflash.github.io/HOIMask/}{\textcolor{brown} {project page}} for more results.
\end{abstract}

\section{Introduction}
\label{sec:intro}
HOI generation has attracted significant attention and plays an important role in a broad range of applications in VR, animation and spatial-intelligence. Text-driven HOI generation \cite{peng2025hoi, cha2024text2hoi, christen2024diffh2o, diller2024cg, zeng2025chainhoi} synthesizes vivid and coherent human-object motions guided by textual descriptions. Achieving realistic HOI requires synthesizing natural human and object motions while deeply understanding the underlying spatio-temporal dynamics between them.

Recently many methods \cite{peng2025hoi, cha2024text2hoi, christen2024diffh2o, diller2024cg, zeng2025chainhoi, li2024controllable, xue2025guiding, wu2025hoi, he2025syncdiff, ji2025onlinehoi} generate HOI motions based on diffusion models \cite{ho2020denoising, songdenoising, dhariwal2021diffusion, rombach2022high, peebles2023scalable}. HOI-Diff \cite{peng2025hoi} introduces a two-stage diffusion framework to generate the contact and motion. CHOIS \cite{li2024controllable} incorporates additional control signals like object waypoints to generate more stable and long-term HOI motions. Recent studies further focus on modeling joint-level interactions \cite{zeng2025chainhoi, huang2025efficient} or designing new interaction representations \cite{xue2025guiding, he2025syncdiff, petrov2025tridi} to improve the quality of generated motions. However, these methods apply diffusion processes to raw HOI motion sequences which involves strong kinematic constraints and tight human–object coupling, where small denoising errors can accumulate across iterative sampling steps, leading to unstable trajectories and physically implausible interactions.

Meanwhile, autoregressive models \cite{vaswani2017attention, parmar2018image, tian2024visual} like generative masked model \cite{chang2022maskgit} further enhance motion generation fidelity by capturing temporal coherence within motion sequences \cite{pinyoanuntapong2024mmm, guo2024momask, pinyoanuntapong2024bamm, pinyoanuntapong2025maskcontrol, meng2025rethinking} compared with diffusion models. Motion autoregressive methods model sequential dependencies by predicting each motion token conditioned on the text token and previously generated tokens. This autoregressive decoding mechanism inherently captures long-range temporal dependencies, resulting in enhanced coherence and generation fidelity, showing great potential for HOI generation. However, due to the inherent complexity of HOI motions, it is essential not only to model the individual temporal dynamics of the human and the object, but also to capture their intricate spatial interaction dependencies.

In this work, we present HOIMask, a novel framework for HOI motion generation based on generative masked modeling in discrete space. First, we propose a 2D HOI VQ-VAE to model HOI motions into discrete 2D human and object token maps using two learned codebooks. In contrast to 1D VQ encodings \cite{huang2025hoigpt} that disregard spatial interactions, our method captures the intricate interaction motion inherent in each HOI sequence. Using 2D convolutions, we generate 2D HOI token map, which captures both human and object motions along with contact information, enabling better spatial awareness. Second, we design a HOI-Mask Transformer to collaboratively model human and object tokens. The HOI-Mask Transformer incorporates spatial-temporal attention block to model detailed motion dependencies, along with human/object-centric attention blocks to capture HOI motions relationships. This architecture allows the model to reason jointly about spatial and temporal dynamics within and across individuals, thereby enhancing its ability to generate interactive HOI motion sequences. Third, to generate more physically plausible interactions, we introduce a novel contact-aware reconstruction guidance during discrete sampling process. We define a discrete probability gradient based on predicted logits, guided by a reconstruction objective at inference stage. The reconstruction objective contains contact loss and penetration loss, and we propose a differentiable Logits Update strategy to optimize the HOI tokens. This guidance enables the HOI-Mask Transformer to optimize logits effectively, resulting in higher-quality and more coherent interactions.

Our main contributions are as follows:
\begin{itemize}
\item We propose HOIMask, the first Vector Quantization (VQ)-based generative masked modeling framework for full-body HOI motion. 
\item  We introduce two key components of HOIMask: (1) a 2D HOI VQ-VAE for capturing complex spatial and temporal interaction to reconstruct high-fidelity HOI motions. (2) an interaction-aware HOI-Mask Transformer for modeling human-object interaction dependencies to generate accurate HOI tokens.
\item We introduce a novel contact-aware reconstruction guidance in discrete sampling space. Specifically, we design a logits update method with differentiable strategy during inference to optimize the HOI tokens which represent more rational HOI motions.  
\item We achieve state-of-the-art performance on FullBodyManipulation \cite{li2023object} and BEHAVE \cite{bhatnagar2022behave} datasets compared to the diffusion-based methods.
\end{itemize}

\section{Related work}

\subsection{Human Object Interaction Generation}
Synthesizing HOI motions has recently gained increasing attention. Extensive datasets \cite{bhatnagar2022behave, li2023object, huang2022intercap, zhao2024m, lu2025humoto} capture HOI motions that includes complex interactions and multiple object categories. GRAB \cite{taheri2020grab} is one of the earliest datasets for full-body human–object interaction. BEHAVE \cite{bhatnagar2022behave} and OMOMO \cite{li2023object} provide more complex interaction scenarios, while IMHD \cite{zhao2024m} contains highly dynamic interactions such as sports activities. Recent works like Text2HOI \cite{cha2024text2hoi} focuses on hand-object interaction generation, while InterDiff \cite{xu2023interdiff} predicts full-body and object motions. HOI-Diff \cite{peng2025hoi} and CG-HOI \cite{diller2024cg} incorporate contact guidance, and CHOIS \cite{li2024controllable} introduces waypoint-based controllability. ChainHOI \cite{zeng2025chainhoi} enhances interaction accuracy through joint-level representations, ROG \cite{xue2025guiding} introduces an Interactive Distance Field (IDF) for interaction modeling, and SyncDiff \cite{he2025syncdiff} presents a frequency-domain modeling paradigm. However, these diffusion-based methods often suffer from error accumulation and inconsistent interactions. In contrast, we present the first VQ-based generative masked model that generates more rational and coherent HOI motions from text.

\subsection{Motion Quantization}
Encoding human motions into discrete tokens has been extensively studied in single-human motion generation \cite{lucas2022posegpt, guo2022tm2t, zhang2023generating, jiang2023motiongpt, lu2024humantomato, zhou2024avatargpt, zou2024parco}. TM2T \cite{guo2022tm2t}, T2M-GPT \cite{zhang2023generating}, and MotionGPT \cite{jiang2023motiongpt} establish mappings between motions and discrete tokens. For HOI representation, HOIGPT \cite{huang2025hoigpt} introduces a dual VQ-VAE \cite{van2017neural} for encoding hand-object motions, but its 1D quantization amplifies interaction errors and simply quantizing HOI motions can't capture the inherent interaction. In this work, we propose a contact-aware HOI quantization that transforms HOI motions and corresponding contact information into discrete 2D token maps, capturing fine-grained spatio-temporal interactions.

\subsection{Generative Masked  Modeling}
Recent studies show that generative masked modeling methods \cite{pinyoanuntapong2024mmm, guo2024momask, pinyoanuntapong2024bamm, meng2025rethinking, jeonghgm3, pinyoanuntapong2025maskcontrol} outperform diffusion-based approaches \cite{chen2023executing, dabral2023mofusion, zhang2023remodiffuse, wang2023fg, zhong2023attt2m, dai2024motionlcm, jin2024local, liu2024bridging, ren2024realistic, zhang2024motiondiffuse, tevethuman, zeng2025light} in generating realistic single-human motions. VQ-based models such as MoMask \cite{guo2024momask} and MMM \cite{pinyoanuntapong2024mmm} employ masked transformers to predict all tokens jointly, while InterMask \cite{javedintermask} extends this to human-human interactions. However, in the HOI generation, diffusion-based methods still dominate. Motivated by these advances, we develop a HOI VQ-VAE and HOI-Mask Transformer to generate high-fidelity, contact-aware human-object interactions.

\section{Methodology}

\begin{figure*}
\centering
\includegraphics[width=1.0\linewidth]{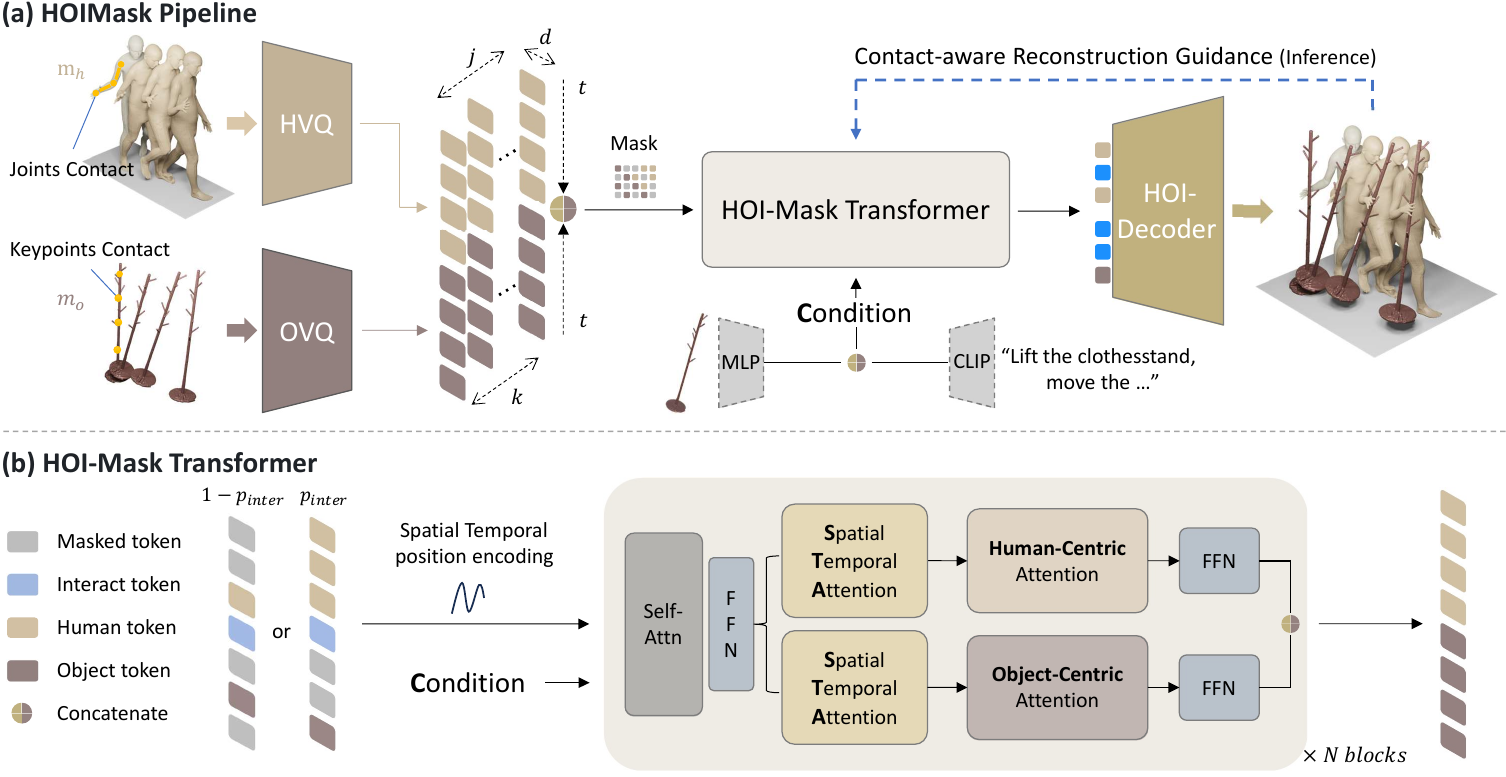}
\caption{Overview of \textbf{HOIMask}. (a) During training, HOI motions first are quantized via HVQ and OVQ into 2D HOI tokens, which are partially masked and predicted by the HOI-Mask Transformer. During inference, the HOI tokens are optimized with contact-aware reconstruction guidance to generate more physically plausible interactions through HOI-Decoder. (b) The HOI-Mask Transformer incorporates spatial-temporal attention and human/object-centric attention to model fine-grained spatio-temporal dependencies.}
\label{fig:overview}
\end{figure*}

\textbf{Overview} Given object geometry and text description, our objective is generating HOI motion. As shown in Figure \ref{fig:overview} (a), we start by learning discrete latent representations via HVQ and OVQ (\ref{sec:HOI VQ-VAE}) to encode human-object motions and their relative contact information into 2D tokens. Then the HOI-Mask Transformer (\ref{sec:HOI-Mask Transformer}) generates 2D HOI tokens conditioned on text and object geometry, which are decoded by the HOI-Decoder to reconstruct HOI motions. During inference, the contact-aware reconstruction guidance (\ref{sec:Guidance}) is applied to optimize the HOI tokens for more physically plausible interactions.

\subsection{HOI VQ-VAE}
\label{sec:HOI VQ-VAE}
\textbf{HVQ} The human motion $M_h \in \mathbb{R}^{T \times J}$ is accompanied by a contact state $C_h \in \mathbb{R}^{T \times J}$, indicating whether each joint is in contact with the object at every time step. Given $M_h$ and $C_h$, a 2D spatio-temporal token map $m_h \in \mathbb{R}^{T \times J \times D}$ is constructed ($T$ is the frames of motion sequence, $J$ is the number of joints, and $D$ is the joint feature dimension), where spatial and temporal dependencies are jointly captured through 2D convolutions. The resulting token maps are then encoded into 2D latent features $\tilde{{l}}_h \in \mathbb{R}^{t \times j \times d}$ (with downsampling ratios $T/t, J/j$), and each $d$-dimensional feature vector is discretized by mapping it to the nearest entry in a learnable human codebook, producing a quantized sequence $l_h = Q(\tilde{{l}}_h) $. \\
\textbf{OVQ} The object motion $M_o \in \mathbb{R}^{T}$ includes the object’s centroid rotation and global 3D position. We uniformly sample $K$ surface points per frame via Poisson Disk Sampling \cite{yuksel2015sample} and pre-calculate their contact states. Similar to the human branch, the motion and contact signals are concatenated into a 2D object token map $m_o \in \mathbb{R}^{T \times K \times D}$ and encoded into latent features $\tilde{{l}}_o \in \mathbb{R}^{t \times k \times d}$. \\
\textbf{HOI-Decoder} The quantized HOI sequence is decoded into motion using a dual-branch 2D HOI-Decoder for humans and objects. The object decoder first generates the object motion conditioned on the object geometry, followed by the human decoder, which generates the human motion conditioned on the object motion and geometry. By conditioning the human motion on the predicted object trajectory and contact states, the generated human movements are constrained to align with the object’s spatial and geometric characteristics, ensuring physically plausible interactions.\\
\textbf{Loss} Following \cite{huang2025hoigpt}, we first incorporate a reconstruction loss and the commitment loss \cite{van2017neural}:
\begin{equation}
\mathcal{L}_{vq} = \lVert {m}_h   - \hat{{m}}_h \rVert_{1} + \lVert {m}_o - \hat{{m}}_o \rVert_{1} + \beta \lVert \tilde{{l}}_h - \mathrm{sg}({l}_h) \rVert_{2}^{2} + \beta \lVert \tilde{{l}}_o - \mathrm{sg}({l}_o) \rVert_{2}^{2},
\end{equation}
where $\mathrm{sg}(\cdot)$ indicates the stop-gradient operation, and $\beta$ serves as a weighting coefficient. To reconstruct both high-fidelity motions and high-quality interactions, we define a training objective composed of a motion loss and an interaction loss.\\
\textbf{Motion Loss} The motions of human and object can be decomposed into global positions $m_p$ and relative rotations $m_r$. The motion loss is defined as the combination of the L1 loss on the predicted positions and the rotation loss relative to the ground truth.
\begin{equation}
\mathcal{L}_{motion} = \lVert m_p - \hat{{m}}_p \rVert_{1} 
+ \lVert {m}_r - \hat{{m}}_r \rVert_{1}.
\end{equation}
\textbf{Interaction Loss} We compute the interaction loss by measuring the spatial relationship between the Poisson Disk Sampled objects $K_o$ and the human pose. Specifically, following \cite{wu2025human}, we compute the object points' positions in the human frame, formulate as:
\begin{equation}
\hat{N}_{h} =  {{R}_K}^{-1} (K_o - T_h),
\end{equation}
where $R_K$ and $T_h$ represent the rotation of the sampled object and the position of the human, respectively. While \cite{wu2025human} focuses solely on hand-level and object-centric interactions, our formulation extends this to full-body and bi-centric modeling, enabling the model to jointly reason about both human and object motions. Furthermore, the human/object-centric contact maps $C_h$ and $C_o$ are used to mask out when the human and object are not in contact. The interaction loss is formulated as:
\begin{equation}
\mathcal{L}_{interact} = \sum_{t=1}^{T} C_h \left\lVert \hat{N}_{h,t} - N_{h,t} \right\rVert_{1} + C_o \left\lVert \hat{N}_{o,t} - N_{o,t} \right\rVert_{1}.
\end{equation}

The overall loss function is a weighted ($\lambda_{\text{motion}}$, $\lambda_{interact}$) sum of motion and interaction loss. The human and object codebooks are jointly updated using Exponential Moving Average (EMA) with periodic resets to ensure stable and consistent quantization, following \cite{zhang2023generating}.
\begin{equation}
\mathcal{L}_{\text{vqvae}} = \mathcal{L}_{vq} + \lambda_{\text{motion}} \mathcal{L}_{motion} + \lambda_{interact} \mathcal{L}_{interact}.
\end{equation}

\subsection{HOI-Mask Transformer}
\label{sec:HOI-Mask Transformer}
Subsequently, the motion tokens of both the human and the object are jointly processed by our HOI-Mask Transformer, which captures rich spatial-temporal dependencies within each entity and across their interactions. An additional interaction token is introduced to explicitly separate human and object tokens, facilitating inter-entity communication. The total input tokens sequence $t^{\prime} \in \mathbb{Z}^{t(j+k)+1}$ is formed by concatenating the flattened 1D human and object tokens with the interaction token. We use the 2D positional encodings \cite{wang2021translating} to impose spatial-temporal structure on the embeddings.\\ 
\textbf{HOI-Mask Mechanism} As shown in Figure \ref{fig:overview} (b), we employ two masking strategies. The first is a random masking strategy, controlled by a cosine scheduling function \cite{chang2022maskgit} $\gamma(\tau) = \cos\!\left(\frac{\pi \tau}{2}\right) \in [0, 1]$, where $\tau \in [0,1]$ that $\tau = 0$ means the sequence is completely corrupted. During random masking, the HOI tokens are randomly masked with ratio $\gamma(\tau)$. To further enhance interaction modeling, we introduce a human-object masking strategy with probability $p_{inter}$. Specifically, the human-object masking randomly masks either the object or the human, and the other one is set to be fully unmasked. This mechanism improves the human and object tokens' dependencies.\\ 
\textbf{STA} In the Transformer module, the 2D HOI tokens are first processed by a self-attention layer \cite{vaswani2017attention} to compute global attention scores. We then decouple the tokens and introduce a Spatial-Temporal Attention (STA) to respectively model the human and object tokens. The STA is composed of two distinct attention mechanisms: spatial attention and temporal attention. In the spatial attention branch, tokens are restricted to attend to other spatial tokens within the same time step. In the temporal attention branch, the same operations will be executed at the same spatial position. The outputs of both branches are finally aggregated through element-wise addition to hybrid spatial and temporal information for more coherent motion representation:
\begin{equation}
{\mathit{STA}} = {Attn}_{(Q_j, K_j, V_j)} + {Attn}_{(Q_t, K_t, V_t)},
\end{equation}
where $Q_j, K_j, V_j$ and $Q_t, K_t, V_t$ are the spatial and temporal queries, keys and values of human and object.\\
\textbf{HOCA}
We further introduce a Human/Object-Centric Attention (HOCA) to enhance cross-entity interaction. The $\mathit{STA}$ features from the human and object branches are fed into the HOCA to exchange information and strengthen the human and object tokens dependencies. In this module, each token of $\mathit{STA}_{human}$ or $\mathit{STA}_{object}$ attends to the entire set of another $\mathit{STA}$ feature, enabling comprehensive modeling of cross-entity dependencies.\\     
\textbf{Training}
The objective of HOI-Mask Transformer is to predict masked tokens conditioned on text and object geometry. Text features are extracted using a frozen CLIP \cite{radford2021learning}, while object features are encoded via an MLP. These are concatenated as the final condition $[c=c_t \vert c_o]$. Given the masked sequence $\tilde{t}$, the model is trained to minimize the negative log-likelihood of the masked tokens, i.e., the cross-entropy loss between predicted probability distribution and one-hot ground truth:
\begin{equation}
\mathcal{L}_{\text{mask}} = \sum_{\tilde{t}_k \in [\text{MASK}]} -\log p_\theta(t_k \mid \tilde{t}, c).
\end{equation}
\textbf{Inference} 
At inference, the HOI-Mask Transformer begins with all positions masked and iteratively generates tokens over $N$ steps. At each step $n$, it predicts token distributions, samples candidates, and re-masks low-confidence tokens $\lceil \gamma(\frac{n}{N} \cdot (t(j+k)+1)) \rceil$ for refinement. This process continues until $n = N$, after which the HOI-Decoder reconstructs motion from the generated 2D HOI tokens. Classifier-free guidance \cite{ho2021classifier} is further employed to improve generation diversity and controllability under conditional and unconditional settings.

\begin{figure*}[h!]
\centering
\includegraphics[width=1.0\linewidth]{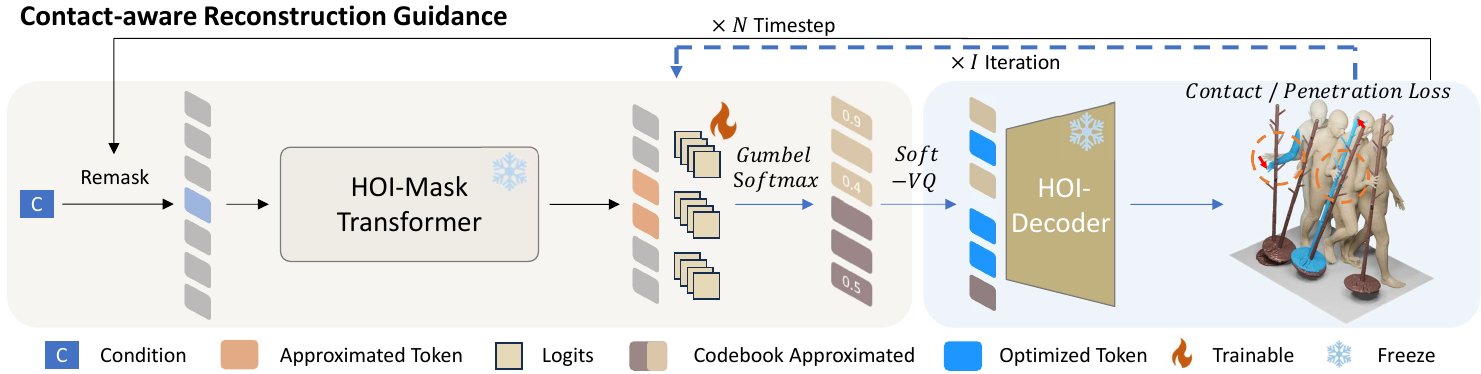}
\caption{At Inference stage, Contact-aware Reconstruction Guidance is applied to optimize HOI tokens. The predicted logits are updated and pass through Soft Vector Quantization (Soft-VQ) to obtain optimized HOI tokens in differentiable paradigm during every iteration.}
\label{fig:inference}
\end{figure*}

\subsection{Contact-aware Reconstruction Guidance}
\label{sec:Guidance}
The HOI-Mask Transformer aims to generate tokens semantically aligned with the given conditions. However, semantic relevance alone does not guarantee physically plausible interactions. Irrational motions may happen because the human-object tokens lack contact-guided coupling between HOI motions. The same problem also occurs in diffusion-based generation models, and some methods \cite{li2024controllable, diller2024cg, wang2026unleashing} design contact guidance during the denoising process to improve interaction quality. Unlike diffusion models that directly perform guidance in continuous space, applying guidance in discrete token space is considerably more challenging due to the non-differentiability of token representations. Therefore, an effective guidance strategy should enable diffusion-like differentiable optimization while effectively refining the human-object tokens. To this end, we introduce a contact-aware reconstruction guidance that incorporates contact signals to fully exploit the model’s potential in HOI generation during inference.

We first formulate two key objectives: a contact loss and a penetration loss. As described in Section \ref{sec:HOI VQ-VAE}, the 2D HOI tokens encode both motion dynamics and contact-aware signals. At every timestep $n$, the HOI-Mask Transformer predicts HOI tokens, which are decoded into motion sequences and corresponding contact maps $\tilde{C_h}$ and $\tilde{C_o}$. These reconstructed maps are used to compute the contact-aware guidance loss, serving as a crucial signal for improving interaction quality.\\
\textbf{Contact Loss} The predicted contact representation includes both human joints and object keypoints contact maps. We apply a thresholding of 0.95 to $\tilde{C_h}$ and $\tilde{C_o}$ to determine whether a contact occurs. Given the SMPL \cite{loper2015smpl} parameters and the generated HOI motions, we compute the 3D positions of the human joints ${J}_h$ and determine which joints are in contact with the object mesh $\tilde{K_o}$, as well as which object regions are contacted by the human throughout the sequence. The contact loss is then formulated as:
\begin{equation}
\mathcal{L}_{c} = \tilde{C_h} \lVert {J}_h - \tilde{{K_o}} \rVert_{1} 
+ \tilde{C_o} \lVert \tilde{{K_o}} - {{J}}_h \rVert_{1}.
\end{equation}
\textbf{Penetration Loss} We design a penetration loss to constrain interpenetration between the human and the object, as well as penetration into the floor. Following \cite{huang2025hoigpt, jiang2021hand, li2024controllable}, for each human joint $J \in \mathcal{J}_{in}^o$ that penetrates the object mesh, we find the closest object vertex and minimize the squared distance. Similar to the human-object penetration, we set a floor height of $(z = 0.02)$ meters to evaluate whether human feet or object keypoints penetrate the floor. The penetration loss is defined as follows:
\begin{equation}
\mathcal{L}_{h_p} = \frac {1} {\mathcal{J}_{in}^o} \sum_{J \in \mathcal{J}_{in}^o} { min \lVert  {J} - \tilde{{K}_o} \rVert_{2}^{2}},
\end{equation}
\begin{equation}
\mathcal{L}_{p} = \lambda_1 \mathcal{L}_{h_p} + \lambda_2 min \lVert  {J}_h^z - z \rVert_{2} + \lambda_3 min \lVert \tilde{{K}_o^z} - z \rVert_{1},
\end{equation}
where $\lambda_1$, $\lambda_2$, $\lambda_3$ denote the loss weights, ${J}_h^z$ and $\tilde{{K}_o^z}$ represent the $z$-coordinates of the human feet and object vertices.\\
\textbf{Logits Update} After calculating the contact and penetration loss, we backpropagate their gradients to the logits generated by the classifier-free guided HOI-Mask Transformer. VQ-based model natively decode the final indices generated from the Transformer process, which is non-differentiable and it is needed for gradient in reconstruction guidance. Recently, \cite{pinyoanuntapong2025maskcontrol} tackle the similar problem using a Differentiable Expectation Sampling (DES). To address this, we introduce a similar differentiable strategy that estimates the probability distribution over codebook indices and applies Gumbel-Softmax \cite{jang2017categorical} sampling to approximate categorical sampling in a differentiable paradigm. Further, we propose a Soft-VQ in HOI-Decoder to transform these updated logits into the HOI tokens.

As shown in Figure \ref{fig:inference}, we set an iteration steps $I$ in every timestep $n$. In timestep $n$, the HOI-Mask Transformer generates the logits, and the optimization process starts in iteration $i$. During training or optimization, motion tokens are sampled from a categorical distribution, a process that is inherently non-differentiable. To make this process differentiable for backpropagation, we apply the Gumbel-Softmax to sample from a categorical distribution. 
\begin{equation}
p_{\theta} \left( x_{k} \mid X_{\overline{\mathbf{M}}}, W, S \right) = \frac{ \exp \left( ( \ell_{k} + g_{k} ) / \tau \right) } { \sum_{j = 1}^{K} \exp \left( ( \ell_{j} + g_{j} ) / \tau \right) },
\end{equation}
where $\ell, \tau$ represent the logits and temperature, $g$ denotes the Gumbel noise with $g_1, ..., g_k$ are independent and identically distributed (i.i.d.) samples from a Gumbel (0, 1) distribution. The logits are passed through the Gumbel-Softmax to obtain a differentiable approximation of discrete token probabilities. The predicted probability $p$ is utilized to reconstruct HOI motion through HOI-Decoder. Contact and penetration loss are applied to provide gradient signals with respect to the logits. Subsequently, the logits at iteration $i{+}1$ are updated using the gradient-based guidance as follows:
\begin{equation}
\mathcal{L}_{i+1} = L_i - \lambda_g \nabla_{L_t} {L_r(l_i, r)},
\end{equation}
where $\lambda_g$ is the guidance scale, ${L_r(l_i, r)}$ represents the gradient of the reconstruction loss $L_r$. The HOI-Decoder uses the predicted probability $p$ to reconstruct HOI motion by querying the discrete token embedding using the index with the highest probability, $argmax_s p_{\theta} (x_s)$, this kind of hard quantization which is also non-differentiable. To obtain continuous and differentiable representations from discrete codebooks, we employ a Soft-VQ module. Unlike standard vector quantization that selects the nearest codeword via a hard assignment, Soft-VQ computes a probabilistic mixture of all codewords based on their probabilities. Formally, given the probability $p_{n,s}$ for the $s$-th codeword in the quantizer, and the corresponding codebook entries $\mathbf{e}_s \in \mathbb{R}^d$, the soft code embedding is obtained as: 
\begin{equation}
\mathbf{z}_{n} = \sum_{S=1}^{S} p_{n,s} \, \mathbf{e}_s.
\end{equation}

\section{Experiments}

\subsection{Dataset}

\textbf{FullBodyManipulation Dataset} We conduct our experiments on the FullBodyManipulation dataset \cite{li2023object}, a comprehensive, large-scale collection that contains 10 hours human-object interaction motion data, accompanied by corresponding videos of 17 subjects interacting with 15 different objects. We follow the train-test split of CHOIS for both training and evaluation. \\
\textbf{BEHAVE Dataset} This dataset \cite{bhatnagar2022behave} contains 20 diverse objects and 8 subjects. We follow the official train-test split and use the version processed by the HOI-Diff \cite{peng2025hoi}.

\begin{table*}[t]
\caption{
Quantitative comparisons on the \textbf{FullBodyManipulation} and \textbf{BEHAVE} datasets.
Bold indicates the best result.
}
\centering
\scriptsize
\setlength{\tabcolsep}{1pt}
\renewcommand{\arraystretch}{1.1}

\begin{tabularx}{\textwidth}{l *{6}{>{\centering\arraybackslash}X}}
\specialrule{1.2pt}{0pt}{0pt}
\multicolumn{7}{c}{\textbf{FullBodyManipulation Dataset}} \\ 

\midrule

Method 
 & FID$^{\scriptstyle\downarrow}$
 & Diversity$^{\scriptstyle\rightarrow}$
 & R-Pre$^{(Top3)}$$^{\scriptstyle\uparrow}$
 & PS$^{\scriptstyle\downarrow}$ 
 & $C_{prec}$$^{\scriptstyle\uparrow}$ 
 & $C_{\%}$ \\

\midrule

GT & $0.01^{\pm 0.000}$ & $1.756^{\pm 0.082}$ & $0.750^{\pm 0.000}$ & $0.060^{\pm 0.000}$ & $-$ & $70.68^{\pm 0.000}$ \\

\rowcolor{gray!10} \multicolumn{7}{l}{\emph{Diffusion Models}}\\
\midrule
 MDM$^{\ast}$ \cite{tevethuman}       & $4.273^{\pm 0.001}$  & $2.391^{\pm 0.082}$ & $0.405^{\pm 0.001}$ & $0.088^{\pm 0.001}$ & $0.39^{\pm 0.007}$ & $22.56^{\pm 0.095}$ \\
InterDiff$^{\ast}$ \cite{xu2023interdiff} & $6.890^{\pm 0.002}$  &  $2.290^{\pm 0.053}$ & $0.605^{\pm 0.000}$ & $0.079^{\pm 0.001}$ & $0.42^{\pm 0.004}$ & $25.63^{\pm 0.100}$ \\
HOI-Diff \cite{peng2025hoi}           & $1.875^{\pm 0.000}$  &  $2.463^{\pm 0.079}$ & $0.792^{\pm 0.000}$ & $0.068^{\pm 0.001}$ & $0.57^{\pm 0.005}$ & $29.20^{\pm 0.090}$\\
CHOIS$^{\ast}$ \cite{li2024controllable}     & $0.810^{\pm 0.001}$  & $2.130^{\pm 0.050}$ & $0.800^{\pm 0.001}$ & $0.066^{\pm 0.002}$ & $0.71^{\pm 0.005}$ & $45.46^{\pm 0.107}$   \\
SemGeoMo$^{\ast}$ \cite{cong2025semgeomo}  & $1.325^{\pm 0.000}$ & $1.453^{\pm 0.057}$ &  $0.778^{\pm 0.002}$ & $0.070^{\pm 0.002}$ & $0.55^{\pm 0.007}$ & $40.25^{\pm 0.090}$  \\
ROG \cite{xue2025guiding}   & $0.199^{\pm 0.002}$   & $1.646^{\pm 0.066}$ & $0.817^{\pm 0.000}$   & $\mathbf{0.054}^{\pm 0.001}$  & $0.47^{\pm 0.008}$ & $27.56^{\pm 0.120}$    \\
LIGHT \cite{wang2026unleashing}  & $\mathbf{0.105}^{\pm 0.002}$  & $\mathbf{1.739}^{\pm 0.070}$  &  $0.779^{\pm 0.000}$ & $0.088^{\pm 0.002}$ & $0.71^{\pm 0.007}$ & $49.57^{\pm 0.091}$ \\

\midrule
\rowcolor{gray!10} \multicolumn{7}{l}{\emph{Generative Masked Model}}\\
\midrule
\textbf{HOIMask}   & ${0.109}^{\pm 0.000}$  & ${1.845}^{\pm 0.083}$  & $\mathbf{0.826}^{\pm 0.000}$ & ${0.065}^{\pm 0.000}$  & $\mathbf{0.74}^{\pm 0.005}$ & $\mathbf{52.96}^{\pm 0.080}$ \\
\end{tabularx}

\begin{tabularx}{\textwidth}{l *{5}{>{\centering\arraybackslash}X}}
\specialrule{1.2pt}{0pt}{0pt}
\multicolumn{6}{c}{\textbf{BEHAVE Dataset}} \\ 
\midrule
Method
 & FID$^{\scriptstyle\downarrow}$
 & Diversity$^{\scriptstyle\rightarrow}$
 & R-Pre$^{(Top3)}$$^{\scriptstyle\uparrow}$ 
 & PS$^{\scriptstyle\downarrow}$ 
 & $C_{\%}$ \\
\midrule

GT & $0.01^{\pm 0.000}$ & $1.896^{\pm 0.047}$ & $0.713^{\pm 0.000}$ & $0.088^{\pm 0.000}$ & $74.20^{\pm 0.000}$ \\

\rowcolor{gray!10} \multicolumn{6}{l}{\emph{Diffusion Models}}\\
\midrule

InterDiff$^{\ast}$ \cite{xu2023interdiff}  &$1.060^{\pm 0.000}$  &$1.697^{\pm 0.097}$ & $0.554^{\pm 0.002}$ & $0.145^{\pm 0.002}$ & $22.59^{\pm 0.113}$  \\
HOI-Diff \cite{peng2025hoi} &$0.725^{\pm 0.000}$  &$1.743^{\pm 0.089}$ & $0.517^{\pm 0.001}$ & $0.178^{\pm 0.001}$ & $36.31^{\pm 0.104}$ \\
CHOIS$^{\ast}$ \cite{li2024controllable} &$0.571^{\pm 0.002}$  &$1.710^{\pm 0.128}$ & $0.606^{\pm 0.002}$ & $0.156^{\pm 0.001}$ & $41.70^{\pm 0.087}$  \\
SemGeoMo$^{\ast}$ \cite{cong2025semgeomo} &$0.985^{\pm 0.001}$  &$1.529^{\pm 0.104}$ & $0.452^{\pm 0.000}$ & $0.165^{\pm 0.002}$ & $43.06^{\pm 0.121}$     \\
ROG \cite{xue2025guiding} &$0.490^{\pm 0.000}$  &$\mathbf{1.887}^{\pm 0.119}$ & $0.595^{\pm 0.002}$ & $0.149^{\pm 0.004}$ & $25.71^{\pm 0.078}$    \\
LIGHT \cite{wang2026unleashing}  & ${0.467}^{\pm 0.005}$  & $1.861^{\pm 0.087}$  &  $0.561^{\pm 0.000}$ & $\mathbf{0.135}^{\pm 0.002}$ & $65.89^{\pm 0.074}$ \\

\midrule
\rowcolor{gray!10} \multicolumn{6}{l}{\emph{Generative Masked Model}}\\

\midrule
\textbf{HOIMask}  & $\mathbf{0.424}^{\pm 0.001}$ &$1.873^{\pm 0.098}$ & $\mathbf{0.606}^{\pm 0.001}$   &${0.141}^{\pm 0.007}$ &$\mathbf{66.90}^{\pm 0.100}$   \\
\specialrule{1.2pt}{0pt}{0pt}
\end{tabularx}

\label{tab:quantitative_all}
\end{table*}

\subsection{Baseline methods}
In our experiment, we compare a series of diffusion-based methods with our VQ-based generative masked model.

\begin{itemize}
\item \textbf{MDM$^{\ast}$} We extend the MDM \cite{tevethuman} by incorporating object motion to enable HOI generation.  
\item \textbf{InterDiff$^{\ast}$} We adapt InterDiff \cite{xu2023interdiff}, a general motion diffusion model, by introducing text conditioning for HOI generation.
\item \textbf{HOI-Diff} \cite{peng2025hoi} employs a dual diffusion model for modeling human and object motions. We follow its official configuration for comparison.
\item \textbf{CHOIS$^{\ast}$} \cite{li2024controllable} introduces additional control signals such as object waypoints. We remove them to align it with our setting.
\item \textbf{SemGeoMo$^{\ast}$} \cite{cong2025semgeomo} takes the object trajectory as additional condition, we remove this conditioning and treat it as the objective to maintain consistency.
\item \textbf{ROG} \cite{xue2025guiding} constructs an Interactive Distance Field (IDF) to generate HOI motions, we keep the official setting to comparison.
\item \textbf{LIGHT} \cite{wang2026unleashing} builds upon diffusion forcing and introduces a data-driven guidance strategy.
\end{itemize}

\subsection{Evaluation Metric}

\textbf{Motion Quality Metric} We first evaluate the motion quality metrics following \cite{guo2022generating, song2024hoianimator}. Fr\'echet Inception Distance (FID) evaluates how closely the generated motions align with the ground truth motions. 
R-Precision (Top3) (R-Pre$^{(Top3)}$), which evaluate semantic alignment between the input text and generated motions. Diversity quantifies the variation within the generated motions.\\
\textbf{Interaction Quality Metric} We validate the interaction quality on Penetration Score (PS), Contact Precision ($C_{prec}$) and Contact Percentage ($C_{\%}$) \cite{li2024controllable}. PS calculates the Signed Distance Function (SDF) values between the human and object meshes in each sequence. $C_{prec}$ measures the accuracy of predicted contact regions. $C_{\%}$ measures the ratio of frames in which human-object interactions are identified as being in contact.

\subsection{Results}

\begin{figure*}[h!]
\centering
\includegraphics[width=1.0\linewidth]{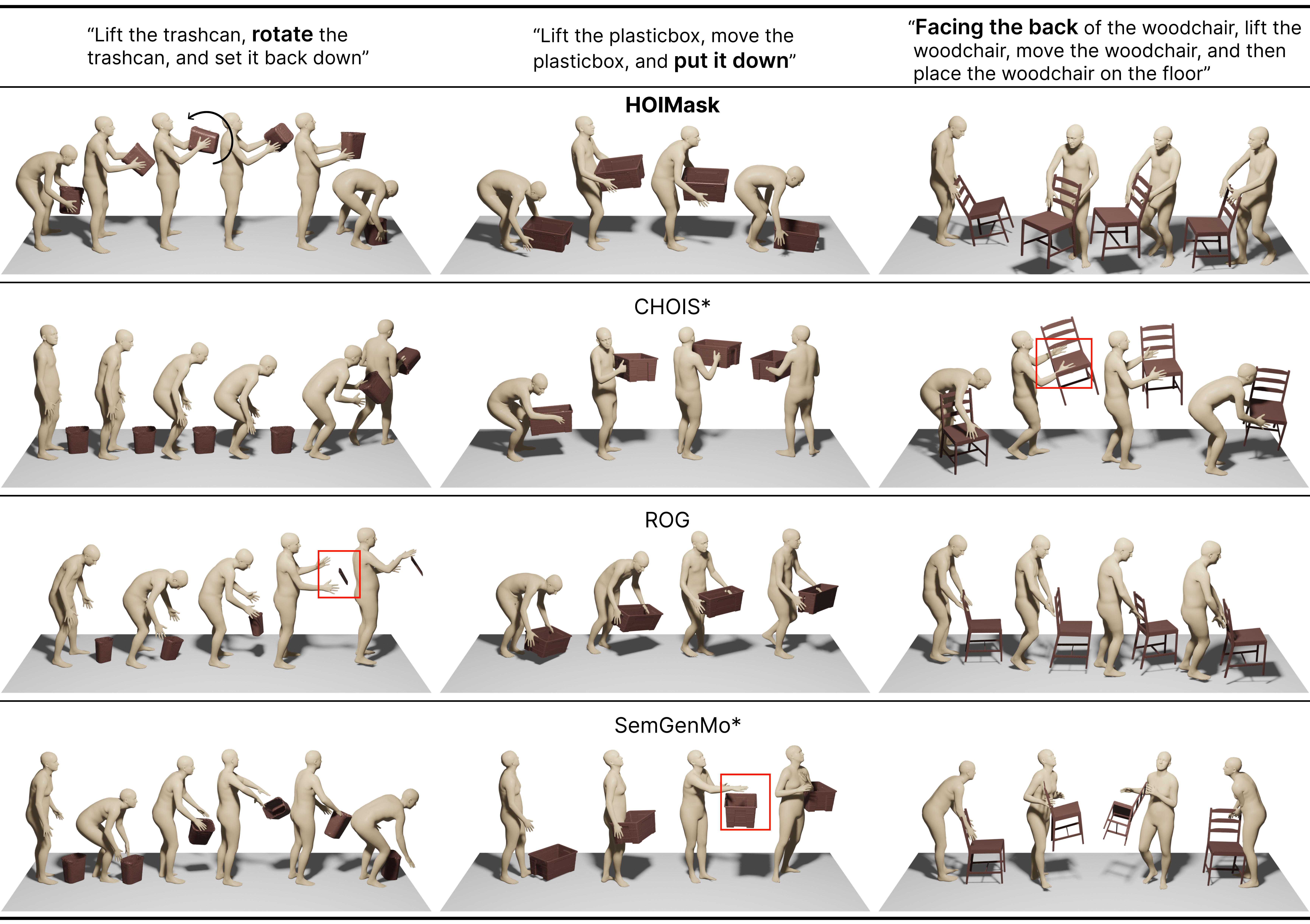}
\caption{Qualitative comparison on the FullBodyManipulation dataset. We use {\color{red}{red boxes}} to highlight the incorrect interaction. The rotating arrow demonstrates the successful action ``rotate''. In contrast, our method generates more realistic and physically consistent human-object interactions than the compared approaches.}
\label{fig:vis}
\end{figure*}

%
\textbf{Quantitative Comparison} Table \ref{tab:quantitative_all} present quantitative comparisons between our \textbf{HOIMask} and various diffusion-based baselines. The results demonstrate that HOIMask achieves state-of-the-art performance, with substantial improvements in FID, R-Pre$^{(Top3)}$, $C_{prec}$ and $C_{\%}$. A lower FID indicates superior realism and visual fidelity of the generated interactions, while the highest R-Precision reflects stronger semantic alignment between motion and textual descriptions. The highest $C_{\%}$ and lowest PS confirm that our model produces more physically plausible and coherent human object interactions.\\
\textbf{Qualitative Comparison} Figure \ref{fig:vis} presents qualitative comparisons between HOIMask, CHOIS \cite{li2024controllable}, ROG \cite{xue2025guiding}, and SemGeoMo \cite{cong2025semgeomo} across three different objects. HOIMask generates motions that are semantically aligned with text inputs and maintain realistic interactions. For the prompt “lift the trashcan, rotate it, and set it back down,” only HOIMask successfully performs the rotation. Overall, HOIMask produces physically plausible and contact-aware interactions, while other methods often show semantic misalignment or penetration artifacts.

\begin{figure}[h!]
\centering
\includegraphics[width=1.0\linewidth]{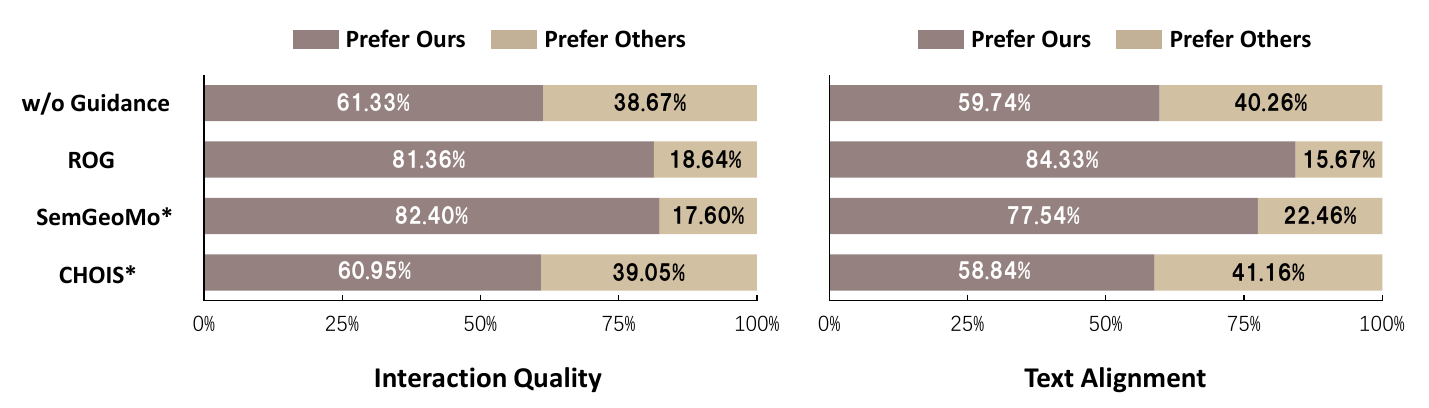}
\caption{User Study. The color bar and numbers represent the preference rate.}
\label{fig:user study}
\end{figure}
\noindent\textbf{User Study} 
We further conduct a user study to validate the generation quality. We randomly generate 20 HOI sequences from our method and a series of diffusion-based methods, and 20 participants were asked to choose the method that better matched the given text description or demonstrated superior interaction quality. As shown in Figure \ref{fig:user study}, HOIMask outperforms other methods in interaction quality and text adherence.

\section{Ablation Studies}
We conduct a series of ablation studies to validate the key components of our HOIMask framework. We first examine the effectiveness of the 2D HOI VQ-VAE design, followed by an analysis of the core modules within the HOI-Mask Transformer. Finally, we evaluate the contribution of the Contact-aware Reconstruction Guidance in enhancing the interaction quality. 

\begin{table*}[htbp]
\caption{Ablation Study results on the FullBodyManipulation test set demonstrating the effectiveness of each key component in the proposed \textbf{HOI-Mask Transformer}. \textbf{Bold} face indicates the best result.}
\centering
\begin{tabularx}{\textwidth}{l *{6}{>{\centering\arraybackslash}X}}
\specialrule{1.2pt}{0pt}{0pt}
Method
 & FID $\downarrow$
 & Diversity $\rightarrow$
 & \mbox{ R-Prec $\uparrow$ } 
 & $C_{prec}$ $\uparrow$ 
 & $C_{\%}$  \\
\midrule
w/o STA   &$0.505^{\pm 0.001}$ & $1.126^{\pm 0.044}$  & $0.781 ^{\pm 0.001}$ & $0.69^{\pm 0.008}$ & $43.92^{\pm 0.056}$  \\
w/o HOCA  & $0.242^{\pm 0.000}$  & $1.496^{\pm 0.098}$    &  $0.794^{\pm 0.000}$   &$0.73^{\pm 0.005}$ &$52.50^{\pm 0.055}$  \\
w/o Guidance    & $0.134^{\pm 0.000}$  & $1.864^{\pm 0.084}$  & $0.800 ^{\pm 0.001}$     &$0.64^{\pm 0.006}$  & $37.54^{\pm 0.067}$   \\
\midrule
\textbf{HOIMask}   &$\mathbf{0.109}^{\pm 0.000}$  & $\mathbf{1.845}^{\pm 0.083}$  & $\mathbf{0.826}^{\pm 0.000}$ & $\mathbf{0.74}^{\pm 0.005}$ & $\mathbf{52.96}^{\pm 0.080}$ \\
\specialrule{1.2pt}{0pt}{0pt}
\end{tabularx}

\label{tab:tab2}
\end{table*}

\subsection{2D or 1D Tokens}
In Table \ref{tab:1_2D token}, we compare our 2D HOI-VQ-VAE with a 1D variant that encodes HOI motion using a 1D convolutional encoder. The results show that whether in motion or interaction quality, the 2D quantization process significantly reduce errors. Moreover, removing either the Interaction Loss or Motion Loss leads to clear performance degradation, highlighting their critical roles in VQ training.    

\begin{table}[th]

\caption{Ablation study results on the FullBodyManipulation test set demonstrating the effectiveness of each key component in the proposed \textbf{HOI VQ-VAE}. \textbf{Bold} face indicates the best result.
}
    
    \centering
    \begin{tabular}{lcccc}
    \specialrule{1.2pt}{0pt}{0pt}
          Method  & MPJPE \(\downarrow\) & $R_o$ \(\downarrow\) & $C_{\%}$  & PS $\downarrow$\\
        \midrule
              GT                    & $0.00$  & $0.00$  & $70.68$  & $0.060$ \\
              1D-VQVAE              & $65.43$ & $0.26$ & $43.15$ & $0.063$\\
              w/o Interaction Loss  & $\mathbf{27.56}$  & $0.10$ & $60.73$ & $0.062$\\
              w/o Motion Loss       & $35.87$  & $0.12$ & $55.41$ & $0.061$\\
              2D-VQVAE              & $28.95$  & $\mathbf{0.08}$ & $\mathbf{65.33}$  & $\mathbf{0.061}$ \\
        \specialrule{1.2pt}{0pt}{0pt}    
    \end{tabular}
    
    \label{tab:1_2D token}
\end{table}

\subsection{HOI-Mask Transformer}
Table \ref{tab:tab2} presents an ablation study to evaluate the key components of HOI-Mask Transformer. The results show that removing the STA, Human/Object Centric Attention or Self Attention degraded the performance.The components of HOI-Mask Transformer can effectively modeling interaction and only connect them together can leverage the model’s maximum potential.


\subsection{Contact-aware Reconstruction Guidance}
The Reconstruction Guidance can significantly improve the interaction quality through logits update process to optimize the HOI tokens. In Table \ref{tab:tab2}, removing this component causes a severe drop in $C_{prec}$ and $C_\%$, highlights this novel technique in generative masked model for improving the interaction quality.
\begin{equation}
\mathcal{L}_{g} = \lambda_c  \mathcal{L}_{c} + \lambda_p  \mathcal{L}_{p},
\end{equation}
where $\lambda_c$ and $\lambda_p$ denote the weights of the contact loss and penetration loss, respectively. We provide additional ablation results on the guidance loss in Table~\ref{tab:guidance_ablation}. Removing the contact loss reduces the contact percentage $C_\%$, while excluding the penetration loss increases the Penetration Score (PS), demonstrating that both terms are essential for Guidance to ensure accurate contact modeling and mitigate physically implausible interpenetration. We further compare our predicted contact guidance with Ground Truth (GT) contact. Although GT contact increases interaction probability, it leads to worse penetration score performance. The GT contact signal may biases the model towards spatial overlap to ensure contact occurrence. Directly applying GT contact as guidance without modeling its coupling with motion dynamic may cause the guidance to drift toward physically implausible configurations. In contrast, the predicted highly correlate contact achieves a better balance between motion fidelity and interaction quality.



\begin{table*}[htbp]
\caption{Ablation study results on \textbf{Contact-aware Reconstruction Guidance}.}
\centering
   
\begin{tabularx}{\textwidth}{l *{6}{>{\centering\arraybackslash}X}}
\specialrule{1.2pt}{0pt}{0pt}
Method
 & FID $\downarrow$
 & Diversity $\rightarrow$
 & \mbox{ R-Prec $\uparrow$ } 
 & $C_{\%}$  
 & PS $\downarrow$ \\
\midrule
              w/o Contact Loss       & $0.121^{\pm 0.001}$ & $1.880^{\pm 0.041}$  &$0.779^{\pm 0.002}$  & $39.01^{\pm 0.082}$ &$\mathbf{0.061}^{\pm 0.000}$  \\
              w/o Penetration Loss   & $0.114^{\pm 0.000}$ & $1.866^{\pm 0.075}$ & $0.814^{\pm 0.002}$  & $50.88^{\pm 0.055}$  &$0.068^{\pm 0.001}$     \\
             w GT Contact   & ${0.112}^{\pm 0.001}$ & $\mathbf{1.823}^{\pm 0.088}$ & $0.821^{\pm 0.001}$  & $\mathbf{54.60}^{\pm 0.049}$  &$0.067^{\pm 0.001}$     \\
              Ours   &$\mathbf{0.109}^{\pm 0.000}$  & ${1.845}^{\pm 0.083}$  & $\mathbf{0.826}^{\pm 0.000}$   & ${52.96}^{\pm 0.080}$ &$0.065^{\pm 0.000}$     \\
        \specialrule{1.2pt}{0pt}{0pt}    
    \end{tabularx}
    
    \label{tab:guidance_ablation}
\end{table*}

\section{Conclusion}
In this paper, we propose HOIMask, the first generative masked modeling framework for HOI generation. HOIMask employs a HOI VQ-VAE to encode HOI motions into 2D discrete token space, preserving fine-grained spatial-temporal structures. Then, a specialized HOI-Mask Transformer is designed to model HOI tokens and capture complex interaction dependencies. Furthermore, we propose a contact-aware reconstruction guidance to optimize HOI tokens during inference, enabling contact signals to guide the generative masked modeling process for producing physically coherent interactions. Extensive experiments demonstrate that HOIMask generates more plausible and semantically consistent HOI motions compared to diffusion-based methods.

\section*{Acknowledgments}
We would like to thank Professor Joshua Zhexue Huang and Professor Yulin He for their help and support in investigation, writing - review \& editing, and funding acquisition. This work was supported by the Science and Technology Major Project of Shenzhen (KJZD20230923114809020), Basic Research Foundation of Shenzhen (JCYJ20250604175602004), and Research Task Assignment Project from Guangdong Laboratory of Artificial Intelligence and Digital Economy (SZ) (No. GML-26420011).

%
%
\bibliographystyle{splncs04}
\bibliography{main}

@String(CVPR  = {IEEE Conf. Comput. Vis. Pattern Recog.})

@String(NeurIPS = {Adv. Neural Inform. Process. Syst.})

@String(AAAI  = {AAAI})

@String(TOG   = {ACM Trans. Graph.})

@String(CVPR  = {CVPR})

@String(NeurIPS = {NeurIPS})

@String(TOG   = {ACM TOG})

@inproceedings{peng2025hoi,
  title={Hoi-diff: Text-driven synthesis of 3d human-object interactions using diffusion models},
  author={Peng, Xiaogang and Xie, Yiming and Wu, Zizhao and Jampani, Varun and Sun, Deqing and Jiang, Huaizu},
  booktitle={Proceedings of the IEEE/CVF Conference on Computer Vision and Pattern Recognition Workshops},
  pages={2878--2888},
  year={2025}
}

@inproceedings{cha2024text2hoi,
  title={Text2hoi: Text-guided 3d motion generation for hand-object interaction},
  author={Cha, Junuk and Kim, Jihyeon and Yoon, Jae Shin and Baek, Seungryul},
  booktitle={Proceedings of the IEEE/CVF Conference on Computer Vision and Pattern Recognition},
  pages={1577--1585},
  year={2024}
}

@inproceedings{christen2024diffh2o,
  title={Diffh2o: Diffusion-based synthesis of hand-object interactions from textual descriptions},
  author={Christen, Sammy and Hampali, Shreyas and Sener, Fadime and Remelli, Edoardo and Hodan, Tomas and Sauser, Eric and Ma, Shugao and Tekin, Bugra},
  booktitle={SIGGRAPH Asia 2024 Conference Papers},
  pages={1--11},
  year={2024}
}

@inproceedings{diller2024cg,
  title={Cg-hoi: Contact-guided 3d human-object interaction generation},
  author={Diller, Christian and Dai, Angela},
  booktitle={Proceedings of the IEEE/CVF Conference on Computer Vision and Pattern Recognition},
  pages={19888--19901},
  year={2024}
}

@inproceedings{zeng2025chainhoi,
  title={Chainhoi: Joint-based kinematic chain modeling for human-object interaction generation},
  author={Zeng, Ling-An and Huang, Guohong and Wei, Yi-Lin and Gu, Shengbo and Tang, Yu-Ming and Meng, Jingke and Zheng, Wei-Shi},
  booktitle={Proceedings of the IEEE/CVF Conference on Computer Vision and Pattern Recognition},
  pages={12358--12369},
  year={2025}
}

@inproceedings{li2024controllable,
  title={Controllable human-object interaction synthesis},
  author={Li, Jiaman and Clegg, Alexander and Mottaghi, Roozbeh and Wu, Jiajun and Puig, Xavier and Liu, C Karen},
  booktitle={European Conference on Computer Vision},
  pages={54--72},
  year={2024}
}

@inproceedings{xue2025guiding,
  title={Guiding Human-Object Interactions with Rich Geometry and Relations},
  author={Xue, Mengqing and Liu, Yifei and Guo, Ling and Huang, Shaoli and Ding, Changxing},
  booktitle={Proceedings of the IEEE/CVF Conference on Computer Vision and Pattern Recognition},
  pages={22714--22723},
  year={2025}
}

@article{wu2025hoi,
  title={HOI-Dyn: Learning Interaction Dynamics for Human-Object Motion Diffusion},
  author={Wu, Lin and Chen, Zhixiang and Lan, Jianglin},
  journal={arXiv preprint arXiv:2507.01737},
  year={2025}
}

@article{ho2020denoising,
  title={Denoising diffusion probabilistic models},
  author={Ho, Jonathan and Jain, Ajay and Abbeel, Pieter},
  journal={Advances in neural information processing systems},
  volume={33},
  pages={6840--6851},
  year={2020}
}

@inproceedings{pinyoanuntapong2024mmm,
  title={Mmm: Generative masked motion model},
  author={Pinyoanuntapong, Ekkasit and Wang, Pu and Lee, Minwoo and Chen, Chen},
  booktitle={Proceedings of the IEEE/CVF Conference on Computer Vision and Pattern Recognition},
  pages={1546--1555},
  year={2024}
}

@inproceedings{guo2024momask,
  title={Momask: Generative masked modeling of 3d human motions},
  author={Guo, Chuan and Mu, Yuxuan and Javed, Muhammad Gohar and Wang, Sen and Cheng, Li},
  booktitle={Proceedings of the IEEE/CVF Conference on Computer Vision and Pattern Recognition},
  pages={1900--1910},
  year={2024}
}

@inproceedings{ji2025onlinehoi,
  title={OnlineHOI: Towards Online Human-Object Interaction Generation and Perception},
  author={Ji, Yihong and Liu, Yunze and Zhuo, Yiyao and Yu, Weijiang and Ma, Fei and Huang, Joshua Zhexue and Yu, Fei},
  booktitle={Proceedings of the 33rd ACM International Conference on Multimedia},
  pages={9395--9403},
  year={2025}
}

@article{huang2025efficient,
  title={Efficient Explicit Joint-level Interaction Modeling with Mamba for Text-guided HOI Generation},
  author={Huang, Guohong and Zeng, Ling-An and Zheng, Zexin and Gu, Shengbo and Zheng, Wei-Shi},
  journal={arXiv preprint arXiv:2503.23121},
  year={2025}
}

@article{li2023object,
  title={Object motion guided human motion synthesis},
  author={Li, Jiaman and Wu, Jiajun and Liu, C Karen},
  journal={ACM Transactions on Graphics (TOG)},
  volume={42},
  number={6},
  pages={1--11},
  year={2023},
  publisher={ACM New York, NY, USA}
}

@inproceedings{chang2022maskgit,
  title={Maskgit: Masked generative image transformer},
  author={Chang, Huiwen and Zhang, Han and Jiang, Lu and Liu, Ce and Freeman, William T},
  booktitle={Proceedings of the IEEE/CVF Conference on Computer Vision and Pattern Recognition},
  pages={11315--11325},
  year={2022}
}

@article{vaswani2017attention,
  title={Attention is all you need},
  author={Vaswani, Ashish and Shazeer, Noam and Parmar, Niki and Uszkoreit, Jakob and Jones, Llion and Gomez, Aidan N and Kaiser, {\L}ukasz and Polosukhin, Illia},
  journal={Advances in neural information processing systems},
  volume={30},
  year={2017}
}

@inproceedings{parmar2018image,
  title={Image transformer},
  author={Parmar, Niki and Vaswani, Ashish and Uszkoreit, Jakob and Kaiser, Lukasz and Shazeer, Noam and Ku, Alexander and Tran, Dustin},
  booktitle={International conference on machine learning},
  pages={4055--4064},
  year={2018},
  organization={PMLR}
}

@article{tian2024visual,
  title={Visual autoregressive modeling: Scalable image generation via next-scale prediction},
  author={Tian, Keyu and Jiang, Yi and Yuan, Zehuan and Peng, Bingyue and Wang, Liwei},
  journal={Advances in neural information processing systems},
  volume={37},
  pages={84839--84865},
  year={2024}
}

@inproceedings{xu2023interdiff,
  title={Interdiff: Generating 3d human-object interactions with physics-informed diffusion},
  author={Xu, Sirui and Li, Zhengyuan and Wang, Yu-Xiong and Gui, Liang-Yan},
  booktitle={Proceedings of the IEEE/CVF International Conference on Computer Vision},
  pages={14928--14940},
  year={2023}
}

@inproceedings{guo2022tm2t,
  title={Tm2t: Stochastic and tokenized modeling for the reciprocal generation of 3d human motions and texts},
  author={Guo, Chuan and Zuo, Xinxin and Wang, Sen and Cheng, Li},
  booktitle={European Conference on Computer Vision},
  pages={580--597},
  year={2022}
}

@inproceedings{zhang2023generating,
  title={Generating human motion from textual descriptions with discrete representations},
  author={Zhang, Jianrong and Zhang, Yangsong and Cun, Xiaodong and Zhang, Yong and Zhao, Hongwei and Lu, Hongtao and Shen, Xi and Shan, Ying},
  booktitle={Proceedings of the IEEE/CVF Conference on Computer Vision and Pattern Recognition},
  pages={14730--14740},
  year={2023}
}

@article{jiang2023motiongpt,
  title={Motiongpt: Human motion as a foreign language},
  author={Jiang, Biao and Chen, Xin and Liu, Wen and Yu, Jingyi and Yu, Gang and Chen, Tao},
  journal={Advances in Neural Information Processing Systems},
  volume={36},
  pages={20067--20079},
  year={2023}
}

@inproceedings{huang2025hoigpt,
  title={HOIGPT: Learning Long-Sequence Hand-Object Interaction with Language Models},
  author={Huang, Mingzhen and Chu, Fu-Jen and Tekin, Bugra and Liang, Kevin J and Ma, Haoyu and Wang, Weiyao and Chen, Xingyu and Gleize, Pierre and Xue, Hongfei and Lyu, Siwei and others},
  booktitle={Proceedings of the IEEE/CVF Conference on Computer Vision and Pattern Recognition},
  pages={7136--7146},
  year={2025}
}

@article{van2017neural,
  title={Neural discrete representation learning},
  author={Van Den Oord, Aaron and Vinyals, Oriol and others},
  journal={Advances in neural information processing systems},
  volume={30},
  year={2017}
}

@inproceedings{javedintermask,
  title={InterMask: 3D Human Interaction Generation via Collaborative Masked Modeling},
  author={Javed, Muhammad Gohar and Li, Xingyu and others},
  booktitle={The Thirteenth International Conference on Learning Representations},
  year={2025}
}

@inproceedings{lucas2022posegpt,
  title={Posegpt: Quantization-based 3d human motion generation and forecasting},
  author={Lucas, Thomas and Baradel, Fabien and Weinzaepfel, Philippe and Rogez, Gr{\'e}gory},
  booktitle={European Conference on Computer Vision},
  pages={417--435},
  year={2022}
}

@inproceedings{yuksel2015sample,
  title={Sample elimination for generating poisson disk sample sets},
  author={Yuksel, Cem},
  booktitle={Computer Graphics Forum},
  volume={34},
  number={2},
  pages={25--32},
  year={2015},
  organization={Wiley Online Library}
}

@inproceedings{wu2025human,
  title={Human-object interaction from human-level instructions},
  author={Wu, Zhen and Li, Jiaman and Xu, Pei and Liu, C Karen},
  booktitle={Proceedings of the IEEE/CVF International Conference on Computer Vision},
  pages={11176--11186},
  year={2025}
}

@article{wang2021translating,
  title={Translating math formula images to LaTeX sequences using deep neural networks with sequence-level training},
  author={Wang, Zelun and Liu, Jyh-Charn},
  journal={International Journal on Document Analysis and Recognition (IJDAR)},
  volume={24},
  number={1},
  pages={63--75},
  year={2021},
  publisher={Springer}
}

@inproceedings{radford2021learning,
  title={Learning transferable visual models from natural language supervision},
  author={Radford, Alec and Kim, Jong Wook and Hallacy, Chris and Ramesh, Aditya and Goh, Gabriel and Agarwal, Sandhini and Sastry, Girish and Askell, Amanda and Mishkin, Pamela and Clark, Jack and others},
  booktitle={Proceedings of the 38st International conference on machine learning},
  pages={8748--8763},
  year={2021},
  organization={PmLR}
}

@inproceedings{ho2021classifier,
  title={Classifier-Free Diffusion Guidance},
  author={Ho, Jonathan and Salimans, Tim},
  booktitle={NeurIPS 2021 Workshop on Deep Generative Models and Downstream Applications},
  year={2021}
}

@article{loper2015smpl,
  title={SMPL: a skinned multi-person linear model},
  author={Loper, Matthew and Mahmood, Naureen and Romero, Javier and Pons-Moll, Gerard and Black, Michael J},
  journal={ACM Transactions on Graphics (TOG)},
  volume={34},
  number={6},
  pages={1--16},
  year={2015},
  publisher={ACM New York, NY, USA}
}

@inproceedings{pinyoanuntapong2025maskcontrol,
  title={MaskControl: Spatio-Temporal Control for Masked Motion Synthesis},
  author={Pinyoanuntapong, Ekkasit and Saleem, Muhammad and Karunratanakul, Korrawe and Wang, Pu and Xue, Hongfei and Chen, Chen and Guo, Chuan and Cao, Junli and Ren, Jian and Tulyakov, Sergey},
  booktitle={Proceedings of the IEEE/CVF International Conference on Computer Vision},
  pages={9955--9965},
  year={2025}
}

@inproceedings{jang2017categorical,
  title={Categorical Reparameterization with Gumbel-Softmax},
  author={Jang, Eric and Gu, Shixiang and Poole, Ben},
  booktitle={The fifth International Conference on Learning Representations},
  year={2017}
}

@inproceedings{tevethuman,
  title={Human Motion Diffusion Model},
  author={Tevet, Guy and Raab, Sigal and Gordon, Brian and Shafir, Yoni and Cohen-or, Daniel and Bermano, Amit Haim},
  booktitle={The Eleventh International Conference on Learning Representations},
  year={2023}
}

@inproceedings{cong2025semgeomo,
  title={Semgeomo: Dynamic contextual human motion generation with semantic and geometric guidance},
  author={Cong, Peishan and Wang, Ziyi and Ma, Yuexin and Yue, Xiangyu},
  booktitle={Proceedings of the IEEE/CVF Conference on Computer Vision and Pattern Recognition},
  pages={17561--17570},
  year={2025}
}

@inproceedings{guo2022generating,
  title={Generating diverse and natural 3d human motions from text},
  author={Guo, Chuan and Zou, Shihao and Zuo, Xinxin and Wang, Sen and Ji, Wei and Li, Xingyu and Cheng, Li},
  booktitle={Proceedings of the IEEE/CVF Conference on Computer Vision and Pattern Recognition},
  pages={5152--5161},
  year={2022}
}

@inproceedings{song2024hoianimator,
  title={Hoianimator: Generating text-prompt human-object animations using novel perceptive diffusion models},
  author={Song, Wenfeng and Zhang, Xinyu and Li, Shuai and Gao, Yang and Hao, Aimin and Hou, Xia and Chen, Chenglizhao and Li, Ning and Qin, Hong},
  booktitle={Proceedings of the IEEE/CVF Conference on Computer Vision and Pattern Recognition},
  pages={811--820},
  year={2024}
}

@inproceedings{meng2025rethinking,
  title={Rethinking Diffusion for Text-Driven Human Motion Generation: Redundant Representations, Evaluation, and Masked Autoregression},
  author={Meng, Zichong and Xie, Yiming and Peng, Xiaogang and Han, Zeyu and Jiang, Huaizu},
  booktitle={Proceedings of the IEEE/CVF Conference on Computer Vision and Pattern Recognition},
  pages={27859--27871},
  year={2025}
}

@inproceedings{songdenoising,
  title={Denoising Diffusion Implicit Models},
  author={Song, Jiaming and Meng, Chenlin and Ermon, Stefano},
  booktitle={The Ninth International Conference on Learning Representations},
  year={2021}
}

@inproceedings{peebles2023scalable,
  title={Scalable diffusion models with transformers},
  author={Peebles, William and Xie, Saining},
  booktitle={Proceedings of the IEEE/CVF International Conference on Computer Vision},
  pages={4195--4205},
  year={2023}
}

@inproceedings{rombach2022high,
  title={High-resolution image synthesis with latent diffusion models},
  author={Rombach, Robin and Blattmann, Andreas and Lorenz, Dominik and Esser, Patrick and Ommer, Bj{\"o}rn},
  booktitle={Proceedings of the IEEE/CVF Conference on Computer Vision and Pattern Recognition},
  pages={10684--10695},
  year={2022}
}

@article{dhariwal2021diffusion,
  title={Diffusion models beat gans on image synthesis},
  author={Dhariwal, Prafulla and Nichol, Alexander},
  journal={Advances in neural information processing systems},
  volume={34},
  pages={8780--8794},
  year={2021}
}

@inproceedings{he2025syncdiff,
  title={Syncdiff: Synchronized motion diffusion for multi-body human-object interaction synthesis},
  author={He, Wenkun and Liu, Yun and Liu, Ruitao and Yi, Li},
  booktitle={Proceedings of the IEEE/CVF International Conference on Computer Vision},
  pages={11731--11743},
  year={2025}
}

@inproceedings{petrov2025tridi,
  title={Tridi: Trilateral diffusion of 3d humans, objects, and interactions},
  author={Petrov, Ilya A and Marin, Riccardo and Chibane, Julian and Pons-Moll, Gerard},
  booktitle={Proceedings of the IEEE/CVF International Conference on Computer Vision},
  pages={5523--5535},
  year={2025}
}

@inproceedings{chen2023executing,
  title={Executing your commands via motion diffusion in latent space},
  author={Chen, Xin and Jiang, Biao and Liu, Wen and Huang, Zilong and Fu, Bin and Chen, Tao and Yu, Gang},
  booktitle={Proceedings of the IEEE/CVF Conference on Computer Vision and Pattern Recognition},
  pages={18000--18010},
  year={2023}
}

@inproceedings{dabral2023mofusion,
  title={Mofusion: A framework for denoising-diffusion-based motion synthesis},
  author={Dabral, Rishabh and Mughal, Muhammad Hamza and Golyanik, Vladislav and Theobalt, Christian},
  booktitle={Proceedings of the IEEE/CVF Conference on Computer Vision and Pattern Recognition},
  pages={9760--9770},
  year={2023}
}

@inproceedings{dai2024motionlcm,
  title={Motionlcm: Real-time controllable motion generation via latent consistency model},
  author={Dai, Wenxun and Chen, Ling-Hao and Wang, Jingbo and Liu, Jinpeng and Dai, Bo and Tang, Yansong},
  booktitle={European Conference on Computer Vision},
  pages={390--408},
  year={2024}

}

@inproceedings{jin2024local,
  title={Local action-guided motion diffusion model for text-to-motion generation},
  author={Jin, Peng and Li, Hao and Cheng, Zesen and Li, Kehan and Yu, Runyi and Liu, Chang and Ji, Xiangyang and Yuan, Li and Chen, Jie},
  booktitle={European Conference on Computer Vision},
  pages={392--409},
  year={2024}
}

@inproceedings{liu2024bridging,
  title={Bridging the gap between human motion and action semantics via kinematic phrases},
  author={Liu, Xinpeng and Li, Yong-Lu and Zeng, Ailing and Zhou, Zizheng and You, Yang and Lu, Cewu},
  booktitle={European Conference on Computer Vision},
  pages={223--240},
  year={2024}
}

@inproceedings{ren2024realistic,
  title={Realistic human motion generation with cross-diffusion models},
  author={Ren, Zeping and Huang, Shaoli and Li, Xiu},
  booktitle={European Conference on Computer Vision},
  pages={345--362},
  year={2024}
}

@inproceedings{wang2023fg,
  title={Fg-t2m: Fine-grained text-driven human motion generation via diffusion model},
  author={Wang, Yin and Leng, Zhiying and Li, Frederick WB and Wu, Shun-Cheng and Liang, Xiaohui},
  booktitle={Proceedings of the IEEE/CVF International Conference on Computer Vision},
  pages={22035--22044},
  year={2023}
}

@inproceedings{zeng2025light,
  title={Light-t2m: A lightweight and fast model for text-to-motion generation},
  author={Zeng, Ling-An and Huang, Guohong and Wu, Gaojie and Zheng, Wei-Shi},
  booktitle={Proceedings of the AAAI Conference on Artificial Intelligence},
  volume={39},
  number={9},
  pages={9797--9805},
  year={2025}
}

@inproceedings{zhang2023remodiffuse,
  title={Remodiffuse: Retrieval-augmented motion diffusion model},
  author={Zhang, Mingyuan and Guo, Xinying and Pan, Liang and Cai, Zhongang and Hong, Fangzhou and Li, Huirong and Yang, Lei and Liu, Ziwei},
  booktitle={Proceedings of the IEEE/CVF International Conference on Computer Vision},
  pages={364--373},
  year={2023}
}

@article{zhang2024motiondiffuse,
  title={Motiondiffuse: Text-driven human motion generation with diffusion model},
  author={Zhang, Mingyuan and Cai, Zhongang and Pan, Liang and Hong, Fangzhou and Guo, Xinying and Yang, Lei and Liu, Ziwei},
  journal={IEEE Transactions on Pattern Analysis and Machine Intelligence},
  volume={46},
  number={6},
  pages={4115--4128},
  year={2024},
  publisher={IEEE}
}

@inproceedings{zhong2023attt2m,
  title={Attt2m: Text-driven human motion generation with multi-perspective attention mechanism},
  author={Zhong, Chongyang and Hu, Lei and Zhang, Zihao and Xia, Shihong},
  booktitle={Proceedings of the IEEE/CVF International Conference on Computer Vision},
  pages={509--519},
  year={2023}
}

@inproceedings{pinyoanuntapong2024bamm,
  title={Bamm: Bidirectional autoregressive motion model},
  author={Pinyoanuntapong, Ekkasit and Saleem, Muhammad Usama and Wang, Pu and Lee, Minwoo and Das, Srijan and Chen, Chen},
  booktitle={European Conference on Computer Vision},
  pages={172--190},
  year={2024}
}

@inproceedings{lu2024humantomato,
  title={HumanTOMATO: text-aligned whole-body motion generation},
  author={Lu, Shunlin and Chen, Ling-Hao and Zeng, Ailing and Lin, Jing and Zhang, Ruimao and Zhang, Lei and Shum, Heung-Yeung},
  booktitle={Proceedings of the 41st International Conference on Machine Learning},
  pages={32939--32977},
  year={2024}
}

@inproceedings{zhou2024avatargpt,
  title={Avatargpt: All-in-one framework for motion understanding planning generation and beyond},
  author={Zhou, Zixiang and Wan, Yu and Wang, Baoyuan},
  booktitle={Proceedings of the IEEE/CVF Conference on Computer Vision and Pattern Recognition},
  pages={1357--1366},
  year={2024}
}

@inproceedings{zou2024parco,
  title={Parco: Part-coordinating text-to-motion synthesis},
  author={Zou, Qiran and Yuan, Shangyuan and Du, Shian and Wang, Yu and Liu, Chang and Xu, Yi and Chen, Jie and Ji, Xiangyang},
  booktitle={European Conference on Computer Vision},
  pages={126--143},
  year={2024}
}

@inproceedings{jiang2021hand,
  title={Hand-object contact consistency reasoning for human grasps generation},
  author={Jiang, Hanwen and Liu, Shaowei and Wang, Jiashun and Wang, Xiaolong},
  booktitle={Proceedings of the IEEE/CVF International Conference on Computer Vision},
  pages={11107--11116},
  year={2021}
}

@inproceedings{jeonghgm3,
  title={HGM$^3$: Hierarchical Generative Masked Motion Modeling with Hard Token Mining},
  author={Jeong, Minjae and Hwang, Yechan and Lee, Jaejin and Jung, Sungyoon and Kim, Won Hwa},
  booktitle={The Thirteenth International Conference on Learning Representations},
  year={2025}
}

@inproceedings{zhou2019continuity,
  title={On the continuity of rotation representations in neural networks},
  author={Zhou, Yi and Barnes, Connelly and Lu, Jingwan and Yang, Jimei and Li, Hao},
  booktitle={Proceedings of the IEEE/CVF Conference on Computer Vision and Pattern Recognition},
  pages={5745--5753},
  year={2019}
}

@inproceedings{xu2025interact,
  title={Interact: Advancing large-scale versatile 3d human-object interaction generation},
  author={Xu, Sirui and others},
  booktitle={CVPR},
  pages={7048--7060},
  year={2025}
}

@inproceedings{bhatnagar2022behave,
  title={Behave: Dataset and method for tracking human object interactions},
  author={Bhatnagar, Bharat Lal and Xie, Xianghui and Petrov, Ilya A and Sminchisescu, Cristian and Theobalt, Christian and Pons-Moll, Gerard},
  booktitle={Proceedings of the IEEE/CVF Conference on Computer Vision and Pattern Recognition},
  pages={15935--15946},
  year={2022}
}

@inproceedings{zhao2024m,
  title={I'm hoi: Inertia-aware monocular capture of 3d human-object interactions},
  author={Zhao, Chengfeng and Zhang, Juze and Du, Jiashen and Shan, Ziwei and Wang, Junye and Yu, Jingyi and Wang, Jingya and Xu, Lan},
  booktitle={Proceedings of the IEEE/CVF Conference on Computer Vision and Pattern Recognition},
  pages={729--741},
  year={2024}
}

@inproceedings{huang2022intercap,
  title={InterCap: Joint markerless 3D tracking of humans and objects in interaction},
  author={Huang, Yinghao and Taheri, Omid and Black, Michael J and Tzionas, Dimitrios},
  booktitle={DAGM German Conference on Pattern Recognition},
  pages={281--299},
  year={2022},
  organization={Springer}
}

@inproceedings{lu2025humoto,
  title={HUMOTO: A 4d dataset of mocap human object interactions},
  author={Lu, Jiaxin and Huang, Chun-Hao Paul and Bhattacharya, Uttaran and Huang, Qixing and Zhou, Yi},
  booktitle={Proceedings of the IEEE/CVF International Conference on Computer Vision},
  pages={10886--10897},
  year={2025}
}

@inproceedings{taheri2020grab,
  title={GRAB: A dataset of whole-body human grasping of objects},
  author={Taheri, Omid and Ghorbani, Nima and Black, Michael J and Tzionas, Dimitrios},
  booktitle={European conference on computer vision},
  pages={581--600},
  year={2020},
  organization={Springer}
}

@inproceedings{wang2026unleashing,
  title={Unleashing guidance without classifiers for human-object interaction animation},
  author={Wang, Ziyin and Xu, Sirui and Guo, Chuan and Zhou, Bing and Gong, Jiangshan and Wang, Jian and Wang, Yu-Xiong and Gui, Liang-Yan},
  booktitle={The Fourteenth International Conference on Learning Representations},
  year={2026}
}

@inproceedings{jiang2023chairs,
  title={Full-body articulated human-object interaction},
  author={Jiang, Nan and Liu, Tengyu and Cao, Zhexuan and Cui, Jieming and Zhang, Zhiyuan and Chen, Yixin and Wang, He and Zhu, Yixin and Huang, Siyuan},
  booktitle={Proceedings of the IEEE/CVF International Conference on Computer Vision},
  pages={9365--9376},
  year={2023}
}

@inproceedings{zhang2023neuraldome,
  title={Neuraldome: A neural modeling pipeline on multi-view human-object interactions},
  author={Zhang, Juze and Luo, Haimin and Yang, Hongdi and Xu, Xinru and Wu, Qianyang and Shi, Ye and Yu, Jingyi and Xu, Lan and Wang, Jingya},
  booktitle={Proceedings of the IEEE/CVF Conference on Computer Vision and Pattern Recognition},
  pages={8834--8845},
  year={2023}
}
\end{document}


\title{Appendix} 


\author{ }
\institute{ }
\authorrunning{Y. Ji et al.}


\maketitle

\section{Overview}
The appendix is structured as follows:
\begin{itemize}
\item Section \ref{sec:Analysis of Diffusion and Generative Masked Modeling}: Analysis of Diffusion and Generative Masked Modeling
\item Section \ref{sec:VQ-VAE Architecture}: HOI VQ-VAE Architecture
\item Section \ref{sec:HOI-Mask Transformer_supp}: Ablation on HOI-Mask Transformer
\item Section \ref{sec:HOI motion and Contact extraction}: Motion and Contact Representation
\item Section \ref{sec:Implementation Details}: Implementation Details
\item Section \ref{sec:Details of Baselines}: Details of Baselines
\item Section \ref{sec:Evaluated on more datasets}: Evaluated on InterAct Datasets
\item Section \ref{sec:Ablation on generation steps and logits iterations}: Ablation on generation steps and logits iterations
\item Section \ref{sec:Inference Speed}: Inference Speed
\item Section \ref{sec:More baselines Visualization}: More baselines Visualization
\item Section \ref{sec:Failure Cases}: Failure Cases
\item Section \ref{sec:Limitations}: Limitations

\end{itemize}

\section{Analysis of Diffusion and Generative Masked Modeling}
\label{sec:Analysis of Diffusion and Generative Masked Modeling}
\subsection{Limiting Diffusion based methods}
Most of diffusion-based methods predict the raw motion data using the DDPM \cite{ho2020denoising} paradigm. The function is defined as:
\begin{equation}
q(x_t \vert x_{t-1}) = \mathcal{N}(\sqrt{\alpha_t}x_{t-1}, (1-\alpha_t) I),
\label{eq1}
\end{equation}
where the constant \({\alpha_t} \in (0,1)\) are hyper-parameters, the \(\{x_t\}_{t=0}^{T}\) denotes the noising sequence, and \(x_{t-1}\) represents the previous step for denoising at the \(t\)-step. However, relying solely on deterministic motion prediction at each timestep may limit training simplicity and reduce sampling diversity. Meanwhile, motion data does not strictly follow a standard normal distribution under conventional z-normalization, due to its heterogeneous structure that combines features from multiple distributions, including 3D continuous variables like joint positions and 6D rotational representations. This heterogeneity introduces dimensional distribution mismatches and complicates the interpolation function in Eq. \ref{eq1} of DDPM. During the forward diffusion process, the different initial distributions of feature groups in $x_0$ cause the corresponding components of $x_t$ to converge toward distinct distributions at timestep $t$, rather than a unified standard distribution. Consequently, initiating the reverse diffusion process from a standard distribution may introduce errors in motion generation. At each timestep, diffusion-based models predict a continuous vector to reconstruct $x_0$ without dimensional alignment, which introduces variability across individual dimensions. As the denoising process iterates, these fluctuations accumulate over time, leading to physically implausible motions. 

\subsection{Benefiting Generative Masked Modeling}
\textbf{VQ-VAE} The redundancy in the data representation facilitates VQ-VAE training and subsequently improves discrete generative modeling \cite{meng2025rethinking}. In VQ-VAE training, the reconstruction loss can be defined as:
\begin{equation}
L_r^{\text{rec}} = \mathcal{L}\big(x_r^{\text{GT}} - x_r^{\text{pred}} + x_{r-o}^{\text{GT}} - x_{r-o}^{\text{pred}}\big) = \mathcal{L}_o^{\text{rec}} + \mathcal{L}\big(x_{r-o}^{\text{GT}} - x_{r-o}^{\text{pred}}\big).
\end{equation}
The $x_r^{\text{GT}}$ and $x_o^{\text{GT}}$ denote the ground truth data with and without redundancy, and the $x_r^{\text{pred}}$ and $x_o^{\text{pred}}$ are the predictions with and without redundancy. The $L_o^{\text{rec}}$ is the reconstruction loss without redundancy, the decomposition shows the redundancy acts as data-level regularization. This implicit regularization reduces model variance, making the learned representations less sensitive to fluctuations in the training data. As a result, it improves generalization and promotes more consistent codebook utilization, leading to a more uniform distribution.

The VQ codebook enforces a discrete one-to-one mapping between tokens and embeddings, ensuring consistent errors across both essential and redundant dimensions, which benefits VQ-based methods under evaluators that consider all dimensions. VQ-VAE establishes a one-to-one correspondence between each discrete code and its embedding. Although this discrete formulation may restrict data diversity, the deterministic mapping constrains outputs to a predefined set of embeddings, yielding stable representations across all dimensions. From a geometric perspective, each codebook embedding corresponds to a Voronoi cell. Consequently, generation errors correspond to deviations of the Voronoi cell centroids, leading to a more uniform error distribution across dimensions and better alignment with evaluators that assess all dimensions.\\
\textbf{Generative Masked Modeling} Generative masked modeling introduces a masking mechanism that is conceptually related to the noise injection process in diffusion models. Instead of gradually corrupting data with Gaussian noise, masked modeling progressively removes partial information by replacing tokens with mask symbols. In this sense, masking can be interpreted as a discrete counterpart to the continuous noise perturbation used in diffusion processes. Compared with diffusion-based generation, generative masked modeling offers several advantages. First, masking operates in a discrete token space, avoiding the need to learn complex continuous denoising trajectories. This often leads to a simpler training objective and improved training stability. Second, masked modeling enables flexible and parallel token prediction, which can significantly reduce generation latency compared to the iterative denoising steps required by diffusion models. Finally, by directly modeling the conditional distribution of missing tokens, masked generative models can better leverage contextual dependencies, resulting in more efficient generation with fewer sampling steps.

\section{HOI VQ-VAE Architecture}
\label{sec:VQ-VAE Architecture}
\begin{figure*}[h!]
\centering
\includegraphics[width=1.0\linewidth]{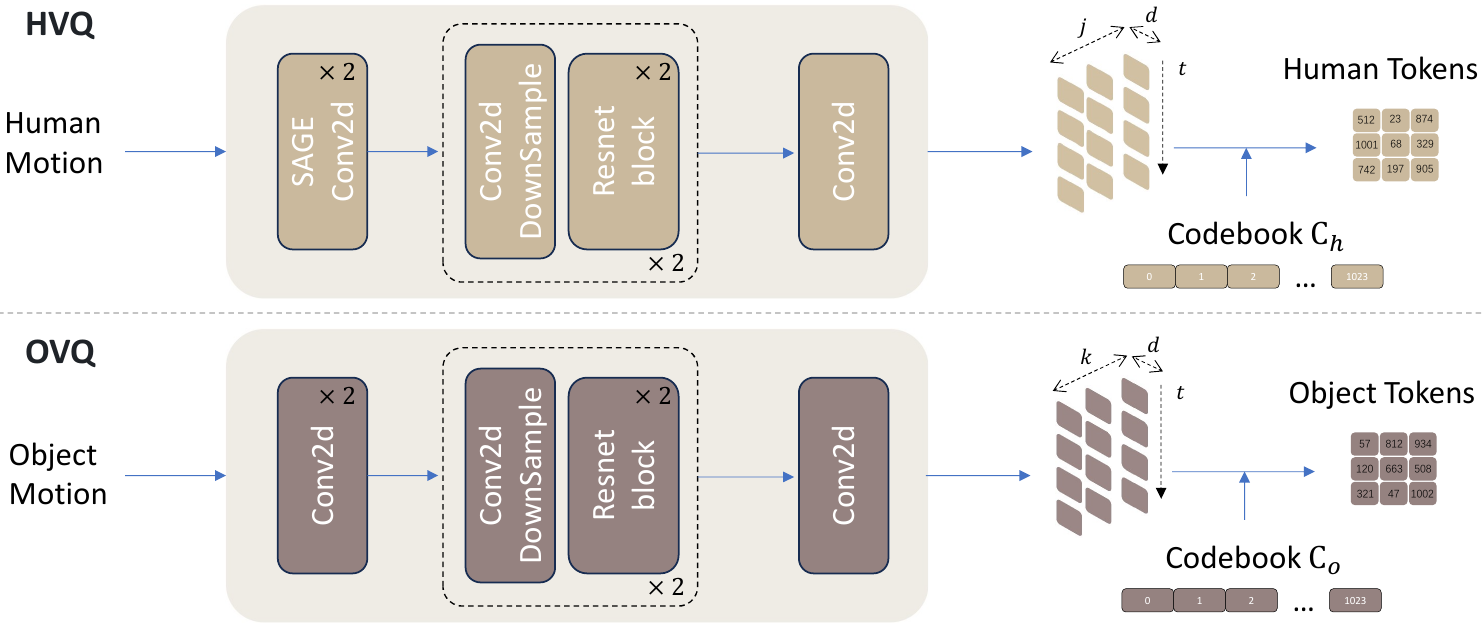}
\caption{2D HOI token map construction. HVQ and OVQ are composed of several 2D convolutional layers, downsamples the input motion sequence from $(T, J, D)$ to $(t, j, d)$ and $(T, K, D)$ to $(t, k, d)$. The resulting latent representation is then quantized by substituting each feature vector with the index of its nearest entry in learned codebook.}
\label{fig:vq}
\end{figure*}

As shown in Figure \ref{fig:vq}, the HOI VQ-VAE employs 2D convolutional residual blocks in both the encoder and decoder. Specifically, we adopt a GraphSAGE-based graph convolutional network (SAGEConv2d) to encode human motion. The human body is modeled as a graph, where nodes correspond to joints and edges follow the kinematic hierarchy. SAGEConv2d aggregates features from neighboring joints to capture spatial dependencies and produces robust joint embeddings. For temporal downsampling, we apply a reduction factor of $t/T = 1/4$, while spatial downsampling differs across modalities, i.e. $j/J = 6/24$ for the human and object branches. The encoder compresses the representation using strided convolutions, and the decoder reconstructs the original resolution through upsampling followed by convolutional layers. The latent space has a dimensionality of 512, and the latent representations are quantized using separate human and object codebooks, each containing $\lvert C_{h/o} \rvert = 1024$ entries. Each feature vector is replaced with the index of its nearest codebook entry, yielding 2D HOI tokens that encode relative spatial–temporal movements. This design preserves the spatial and temporal integrity of the motion data, thereby enhancing the model’s capacity to generate realistic and context-aware human–object interactions. 

\section{Ablation on HOI-Mask Transformer}
\label{sec:HOI-Mask Transformer_supp}
\subsection{Ablation on Human-Object Masking}
In the HOI-Mask mechanism, two types of masking strategies are employed: random masking and human–object masking. Random masking encourages the model to predict randomly selected tokens from both entities, while human–object masking—where tokens from either the human or the object are masked, while the other is fully unmasked, focuses on capturing cross-entity dependencies that are essential for generating coherent interactions. During training, the model alternates between these two strategies according to the masking probability parameters: human–object masking is applied with probability $p_{inter}$, and random masking with probability $(1 - p_{inter})$. To investigate the impact of $p_{inter}$, we experimented with values of $(0.1, 0.6)$ and found that $p_{inter} = 0.2$ achieves the best trade-off between interaction understanding and generation performance, as shown in Table \ref{tab:p_inter}.

\begin{table}[th]
    \caption{Ablation study results on probability of human-object masking. The gray line represents the configuration used in HOIMask.}
    \centering
    
    \begin{tabular}{cccccc}
    \specialrule{1.2pt}{0pt}{0pt}
         & $p_{inter}$  & FID \(\downarrow\) & R-Prec (Top3) \(\uparrow\)  & PS \(\downarrow\) & $C_{\%}$\\
        \hline
            & 0.6       & $0.1278$   & $0.8570$    &$0.070$ &$59.40$  \\
            & 0.5       & $0.1106$   & $0.8433$    &$0.067$ &$56.36$  \\
            & 0.4       & $0.1020$   & $0.8125$    &$0.065$ &$53.43$  \\
            & 0.3       & $0.1260$   & $0.8196$     &$0.066$ &$59.36$ \\
          \rowcolor{gray!20} & 0.2   &  $0.1091$  & $0.8264$ & $0.065$ & $52.96$ \\
            & 0.1      & $0.1278$    & $0.8188$   &$0.070$ &$52.37$     \\
        \specialrule{1.2pt}{0pt}{0pt}    
    \end{tabular}
    
    \label{tab:p_inter}
\end{table}

\subsection{Ablation on Key Components}
\begin{figure}[h!]
\centering
\includegraphics[width=0.8\linewidth]{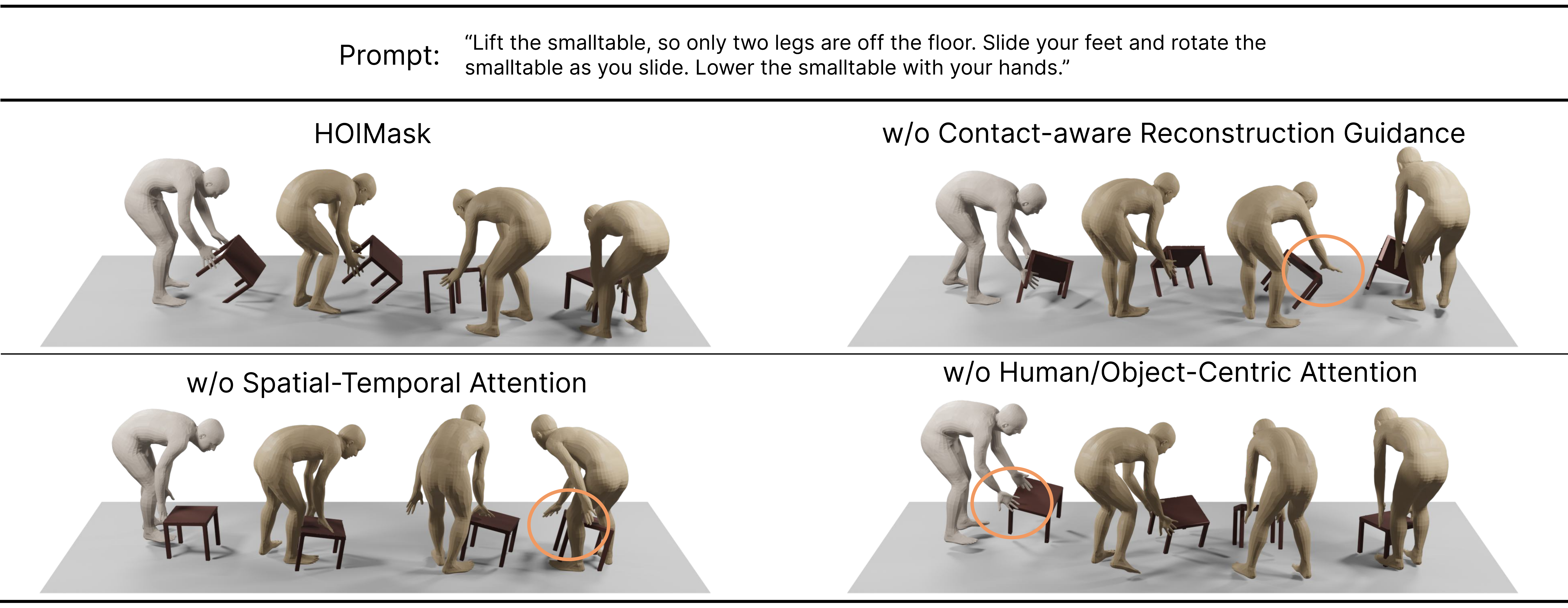}
\caption{Qualitative results for the ablation study on HOIMask. Removing specific modules leads to a noticeable degradation in the quality of the generated results.}
\label{fig:abla}
\end{figure}

As shown in Figure \ref{fig:abla}, we provide qualitative results to evaluate the key components of HOIMask. Without Contact-aware Reconstruction Guidance, STA or Human/Object-Centric Attention may lead to implausible interaction and inconsistent HOI motions. 

\section{Motion and Contact Representation}
\label{sec:HOI motion and Contact extraction}
\textbf{Motion Representation} For human motion representation, we utilize the SMPL model \cite{loper2015smpl}, which includes parameters for the head and body. At each time step $t$, the human pose state is represented by the global joint rotation $Q_t \in \mathbb{R}^{22\times6}$ and the global joint positions $J_t \in \mathbb{R}^{24\times3}$, where the rotations are expressed using the 6D continuous representation \cite{zhou2019continuity}. For object motion, we adopt the global object position $\mathbf{Z}_t \in \mathbb{R}^{N\times3}$ and the relative rotation $\mathbf{Z}_r \in \mathbb{R}^{N\times9}$ to form the combined pose representation $\mathbf{Z} = [\mathbf{Z}_t \mid \mathbf{Z}_r]$, which encodes both translational and rotational states of the object.\\
\textbf{Contact Representation} We directly extract human joint positions from the SMPL parameters. Specifically, 24 human joints are sampled from the SMPL mesh, which consists of 6,890 vertices. The distances between each human joint and the object vertices are computed, and a distance threshold of $0.05 m$ is applied to determine the contact labels. The raw object point clouds typically contain a large number of points, and different object categories vary significantly in vertex density, leading to high computational cost and inconsistent input dimensionality during training. To address this issue and obtain a unified representation across different objects, we apply Poisson Disk Sampling to uniformly downsample each object to 60 surface points. The sampled points are then transformed for every motion sequence according to the corresponding object rotation and translation. Similar to the human contact computation, we calculate the distances between sampled object points and human joints at each time step to extract the object contact regions. The resulting object contact features are reshaped as $\mathbf{C_o} \in \mathbb{R}^{N\times20\times3}$ and concatenated with the object motion features $\mathbf{Z'} \in \mathbb{R}^{N\times4\times3}$, which are derived by reshaping $\mathbf{Z} \in \mathbb{R}^{N\times12}$. This composition forms 24 object nodes aligned with the dimensionality of human joints. Finally, the human and object contact features are fused to construct the HOI motion representation.


\section{Implementation Details}
\label{sec:Implementation Details}
\textbf{HOI VQ-VAE} The HOI VQ-VAE uses separate human and object codebooks, each containing 1024 entries with a latent dimensionality of 512. The model is trained for 200 epochs with a batch size of 256 on an NVIDIA A100 (40GB) GPU. The learning rate is initialized at 0.0002 and follows a multi-step decay schedule, decreasing by a factor of 0.1 at 70$\%$ and 85$\%$ of the total training iterations. A linear warm-up is applied during the first 25$\%$ of training. The commitment loss weight $\beta$, motion loss weight $\lambda_{\text{motion}}$, and interaction loss weight $\lambda_{\text{interact}}$ are set to 0.02, 0.01, and 1, respectively.\\
\textbf{HOI-Mask Transformer} For training the HOI-Mask Transformer, the HOI VQ-VAE is frozen and used to quantize HOI motion into 2D HOI tokens. We adopt the CLIP ViT-L/14@336px model to extract text embeddings. Object geometry is represented using the Basis Point Set (BPS). The Transformer is trained to generate HOI tokens conditioned on text and object features. Training is performed for 200 epochs with a batch size of 64 on an NVIDIA A100 (40GB) GPU. A condition drop probability of 0.1 is applied, enabling the model to learn effectively under both text-conditioned and unconditional settings.\\
\textbf{Inference} During inference, both the HOI VQ-VAE and HOI-Mask Transformer are frozen. The Transformer is run for 20 generation steps with a Classifier-free guidance (CFG) scale of 2 and a sampling temperature of 1, balancing diversity and consistency in the generated motion.
For Contact-aware Reconstruction Guidance, the logits are updated using a reconstruction loss with a learning rate of 0.01 over 20 optimization iterations at each generation step.

\begin{figure*}[h!]
\centering
\includegraphics[width=1.0\linewidth]{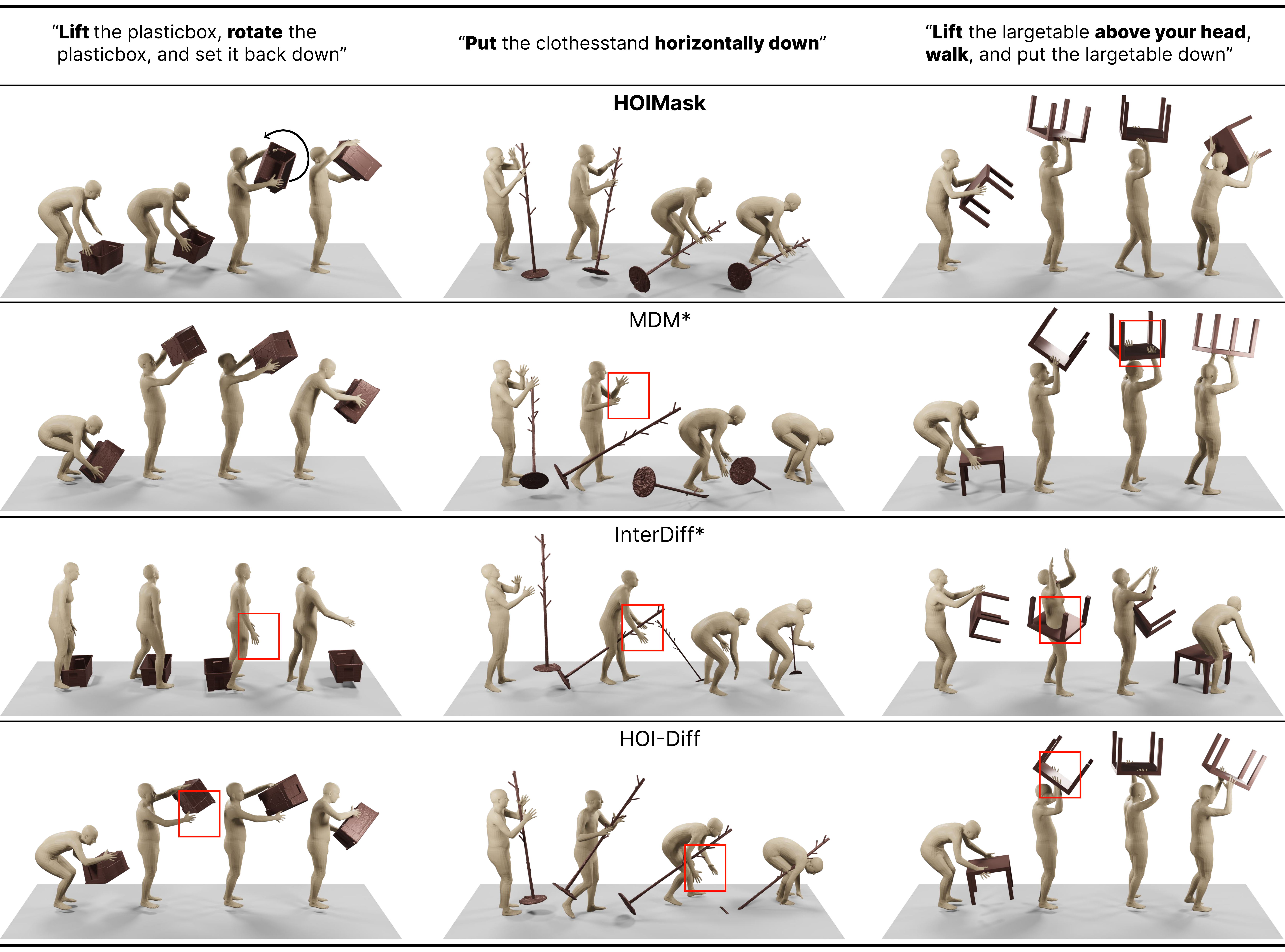}
\caption{Qualitative comparisons on more baseline methods. We use \textcolor{red} {red boxes} to highlight the incorrect interaction. The Rotating arrow demonstrates the successful action "rotate".  In contrast, our method generates more realistic and physically consistent human–object interactions than the compared approaches.}
\label{fig:vis more baseline}
\end{figure*}

\section{Details of Baselines}
\label{sec:Details of Baselines}
We provide more details about the modified methods.\\
\textbf{MDM$^{\ast}$} \cite{tevethuman} MDM is a pioneering diffusion-based approach for human motion generation. To adapt it to HOI generation, we concatenate the human and object motion sequences to form a unified motion representation, enabling the model to operate in the HOI setting.\\
\textbf{InterDiff$^{\ast}$} \cite{xu2023interdiff} InterDiff is originally designed for HOI prediction conditioned on past HOI trajectories. To support text-driven HOI generation, we replace the historical motion inputs with textual descriptions. Specifically, we adjust the feature dimensionality of the model and employ the CLIP text encoder to extract text-driven representations.\\
\textbf{CHOIS$^{\ast}$} \cite{li2024controllable} CHOIS introduces additional control signals like waypoints and initial/end positions to guide the diffusion process. Under our setting, which conditions only on text and object geometry, we remove the initial human–object states and waypoint controls to align CHOIS with our input modality. During inference, we follow the official configuration and apply the same contact-guidance procedures for fair comparison.\\
\textbf{SemGeoMo$^{\ast}$} \cite{cong2025semgeomo} SemGeoMo incorporates semantic and contact prediction modules to guide the diffusion model toward realistic human motion generation. While the original method takes object trajectories and fine-grained text as inputs, we remove the object trajectory and treat it as the prediction target to match our HOI generation setting.\\

\begin{figure*}[h!]
\centering
\includegraphics[width=1.0\linewidth]{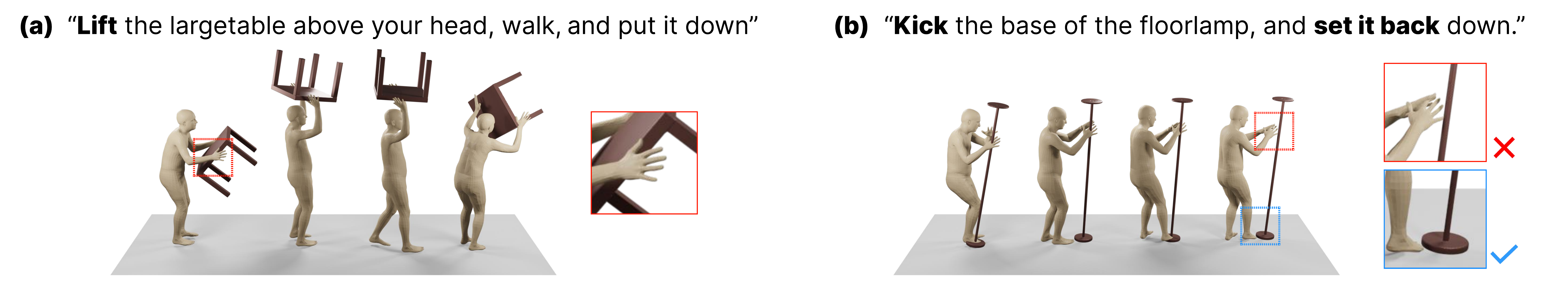}
\caption{Visualization of Failure Cases. We use \textcolor{red}{red box} to highlight the incorrect interaction, the \textcolor[RGB]{47, 154, 255}{blue box} indicates the correct interaction.}
\label{fig:failure_vis}
\end{figure*}

\section{Evaluated on InterAct Dataset}
\label{sec:Evaluated on more datasets}
We conduct preliminary experiments on the InterAct~\cite{xu2025interact} dataset, which consists of the OMOMO \cite{li2023object}, NeuralDome \cite{zhang2023neuraldome}, IMHD \cite{zhao2024m}, and Chairs \cite{jiang2023chairs} datasets. Low-quality interaction sequences are filtered out to construct a large-scale and unified motion representation dataset. We train and evaluate our model following the official data split protocol of LIGHT \cite{wang2026unleashing}. The quantitative results are reported in Table~\ref{tab: interact}.

\begin{table*}[htbp]
    \centering
    \caption{Performance on InterAct \cite{xu2025interact} dataset.}
    \renewcommand{\arraystretch}{1.4}
    \small
    \resizebox{\linewidth}{!}{
    \begin{tabular}{cccccccc}
    \specialrule{1.2pt}{0pt}{0pt}
         & Model  & FID \(\downarrow\) & Diversity & R-Prec (Top3) \(\uparrow\) & PS \(\downarrow\) & $C_{prec}$ \(\uparrow\) & $C_\%$  \\
        \hline
            & HOI-Diff \cite{peng2025hoi}            & $0.689$  &  $7.620$ & $0.740$  & $0.103$ & $0.722$ &$0.084$                \\
            & CHOIS \cite{li2024controllable}           & $0.572$   & $7.717$ & $0.766$ & $0.131$ & $0.710$ & $0.115$            \\
            & LIGHT \cite{wang2026unleashing}             & $0.148$   & $7.712$ & $0.754$ & $0.132$     &$0.731$ & $0.132$              \\
            & HOIMask        &  $0.139$ &  $7.790$  & $0.769$ & $0.132$ & $0.750$  & $0.136$  \\
        \specialrule{1.2pt}{0pt}{0pt}    
    \end{tabular}}
    
    \label{tab: interact}
\end{table*}

\section{Ablation on generation steps and logits iterations}
\label{sec:Ablation on generation steps and logits iterations}
In the HOI-Mask Transformer, we set the number of generation steps to 20, during which the model iteratively generates tokens through a mask–remask process. As shown in Table \ref{tab: generation step}, we conduct an ablation study to analyze the effect of the number of generation steps. The results show that using fewer steps leads to a clear degradation in performance—for example, a significant increase in FID—indicating that the masking mechanism requires sufficiently many iterative refinements to progressively decompose the generation process into manageable sub-tasks. With too few steps, the model lacks the iterative capacity needed to resolve ambiguous or uncertain regions, resulting in poorer motion quality and weaker human–object interactions.
\begin{table*}[htbp]
\centering
\caption{Quantitative results on generation steps. \textbf{Bold} face indicates the best result. The gray line represents the configuration used in HOIMask.}
\renewcommand{\arraystretch}{1.4}
\begin{tabularx}{\textwidth}{c *{5}{>{\centering\arraybackslash}X}}
\specialrule{1.2pt}{0pt}{0pt}

Generation Steps
 & FID$\downarrow$
 & Diversity$\rightarrow$
 & \mbox {R-Prec(Top3)$\uparrow$} 
 & $C_{prec}$$\uparrow$ 
 & $C_{\%}$ \\ 
GT       & $0.01^{\pm 0.000}$ & $1.7557^{\pm 0.082}$ & $0.7504^{\pm 0.000}$ & $-$ & $70.68^{\pm 0.000}$   \\
\midrule

5             &$0.3409^{\pm 0.001}$  & $2.0731^{\pm 0.097}$  & $0.8212^{\pm 0.000}$  &${0.70}^{\pm 0.008}$   & ${33.15}^{\pm 0.124}$ \\
10            &$0.2349^{\pm 0.001}$ & $2.0135^{\pm 0.087}$  &  $0.8037^{\pm 0.001}$ &$0.74^{\pm 0.003}$ & $47.55^{\pm 0.092}$ \\
\rowcolor{gray!20} 20      & $\mathbf{0.1091}^{\pm 0.000}$  & $\mathbf{1.8450}^{\pm 0.083}$  & $\mathbf{0.8264}^{\pm 0.000}$  & $\mathbf{0.74}^{\pm 0.005}$ & ${52.96}^{\pm 0.080}$   \\
40   & $0.1469^{\pm 0.000}$  & $1.9128^{\pm 0.092}$ & $0.8152^{\pm 0.001}$  &  $0.73^{\pm 0.004}$  & $\mathbf{54.98}^{\pm 0.110}$   \\
\specialrule{1.2pt}{0pt}{0pt}
\end{tabularx}

\label{tab: generation step}
\end{table*}

For Contact-aware Reconstruction Guidance, we set the number of optimization iterations for logits refinement to 20 at each generation step. As shown in Table \ref{tab: logits update inter}, we further perform an ablation study on the number of logits-update iterations. The results demonstrate that using more than 20 iterations tends to cause overfitting, where the HOI tokens become trapped in overly aggressive negative optimization, limiting diversity and leading to unnatural or collapsed interactions. Conversely, too few iterations weaken the effect of contact-aware guidance, reducing contact accuracy and physical plausibility. Overall, 20 iterations strike a good balance between reconstruction quality, stable optimization, and interaction realism.
 
\begin{table*}[htbp]
\centering
\caption{Quantitative results on logits update iterations. \textbf{Bold} face indicates the best result. The gray line represents the configuration used in HOIMask.}
\renewcommand{\arraystretch}{1.4}
\begin{tabularx}{\textwidth}{c *{5}{>{\centering\arraybackslash}X}}
\specialrule{1.2pt}{0pt}{0pt}

Update Iterations 
 & FID$\downarrow$
 & Diversity$\rightarrow$
 & \mbox {R-Prec(Top3)$\uparrow$} 
 & $C_{prec}$$\uparrow$ 
 & $C_{\%}$ \\ 
GT       & $0.01^{\pm 0.000}$ & $1.7557^{\pm 0.082}$ & $0.7504^{\pm 0.000}$ & $-$ & $70.68^{\pm 0.000}$   \\
\midrule
10        & $0.1117^{\pm 0.001}$  & $\mathbf{1.8231}^{\pm 0.1255}$   & $0.7908^{\pm 0.0000}$ &$0.72^{\pm 0.007}$  & $49.80^{\pm 0.120}$ \\
\rowcolor{gray!20} 20  & $\mathbf{0.1091}^{\pm 0.000}$  & ${1.8450}^{\pm 0.083}$  & $\mathbf{0.8264}^{\pm 0.000}$  & ${0.74}^{\pm 0.005}$ & ${52.96}^{\pm 0.080}$ \\
40      & $0.1613^{\pm 0.0000}$  & $1.9414^{\pm 0.0885}$  & $0.7908^{\pm 0.0001}$   &$\mathbf{0.75}^{\pm 0.004}$ & $\mathbf{53.93}^{\pm 0.098}$   \\
80   & $0.2093^{\pm 0.0001}$ & $1.2006^{\pm 0.1090}$ & $0.7714^{\pm 0.0001}$  &$0.74^{\pm 0.008}$  & $53.79^{\pm 0.105}$  \\
\specialrule{1.2pt}{0pt}{0pt}
\end{tabularx}

\label{tab: logits update inter}
\end{table*}

\section{Inference Speed}
\label{sec:Inference Speed}
We compare the inference speed of HOIMask under two configurations—with and without Contact-aware Reconstruction Guidance—against state-of-the-art diffusion-based methods. We measure the time required to generate a single HOI motion sequence using the same prompt and sequence length across all methods. As shown in Table \ref{tab:speed}, although applying Contact-aware Reconstruction Guidance significantly increases the inference time ($55.80  s$ vs. $4.57  s$), the overall runtime remains comparable to diffusion-based methods such as ROG and SemGeoMo, which exhibit similar time costs.

\begin{table}[h!]
    \caption{Results on inference speed.}
    \centering
    
    \begin{tabular}{llllll}
    \specialrule{1.2pt}{0pt}{0pt}
         & Model  & Speed\(\downarrow\) \\
        \hline
             & MDM$^{\ast}$ \cite{tevethuman}                  &  $4.91  s$    &    \\
             & CHOIS$^{\ast}$ \cite{li2024controllable}                &  $7.89  s$   &        \\
             & ROG \cite{xue2025guiding}                &  $42.04  s$ \\
             & SemGeoMo$^{\ast}$ \cite{cong2025semgeomo}              &  $221.48  s$     \\
             & HOIMask (w/o Guidance)   &  $ \mathbf{ 4.57 } s$         \\
             & HOIMask                   &  $55.80  s$   &         \\
             
        \specialrule{1.2pt}{0pt}{0pt}    
    \end{tabular}
    
    \label{tab:speed}
\end{table}

\section{More baselines Visualization}
\label{sec:More baselines Visualization}
In this section, we present additional qualitative comparisons against more baseline methods, including HOI-Diff, InterDiff$^{\ast}$ and MDM$^{\ast}$. As shown in Figure \ref{fig:vis more baseline}, these diffusion-based methods often produce implausible or physically inconsistent human–object interactions, further demonstrating the ability of our approach to generate high-fidelity and realistic interaction motions.

\section{Failure Cases}
\label{sec:Failure Cases}
As shown in Figure \ref{fig:failure_vis}, we present two types of failure cases produced by our method. In case (a), since the human motions in the FullBodyManipulation dataset \cite{li2023object} are based on the SMPL model \cite{loper2015smpl}, which does not provide explicit finger-joint representations, HOIMask cannot accurately model fine-grained finger articulations. This occasionally results in unintended penetrations between the fingers and the manipulated objects. In case (b), for more complex interactions that require coordinated control of both the hands and feet, our model may struggle to handle all involved body parts simultaneously. 

\section{Limitations}
\label{sec:Limitations}
Although HOIMask is capable of generating realistic HOI motions from textual descriptions, it still exhibits several limitations. First, because the FullBodyManipulation dataset \cite{li2023object} is based on the SMPL model \cite{loper2015smpl}, which does not include detailed finger articulations, HOIMask cannot accurately model fine-grained finger motions, leading to occasional intersections between fingers and objects. Second, as discussed in Section \ref{sec:Inference Speed}, there exists a performance trade-off between generation quality and inference speed, particularly when applying Contact-aware Reconstruction Guidance.

\bibliographystyle{splncs04}
\bibliography{main}